\documentclass[sn-mathphys]{sn-jnl}% Math and Physical Sciences Numbered Reference Style 
\usepackage{graphicx}%
\usepackage{multirow}%
\usepackage{amsmath,amssymb,amsfonts}%
\usepackage{amsthm}%
\usepackage{mathrsfs}%
\usepackage[dvipsnames]{xcolor}
\usepackage{booktabs}%
\usepackage{pgfplots}
\usepackage{lscape}
\usepackage{tabularx}
\usepackage{arydshln} % Package for dashed lines
\usepackage[linesnumbered,lined,boxed,commentsnumbered]{algorithm2e}
\pgfplotsset{compat=1.18}
\begin{document}

\title{Deceiving Question-Answering Models: A Hybrid Word-Level Adversarial Approach}

\author[1]{\fnm{Jiyao} \sur{Li}}\email{jiyao.li-1@student.uts.edu.au}

\author[1]{\fnm{Mingze} \sur{Ni}}\email{mingze.ni@uts.edu.au}
% \author[1]{\fnm{Yifei} \sur{Dong}}\email{yifei.dong@uts.edu.au}

% \author[2]{\fnm{Tianqing} \sur{Zhu}}\email{tqzhu@cityu.edu.mo}
\author[2]{\fnm{Yongshun} \sur{Gong}}\email{ysgong@sdu.edu.cn}

\author*[1]{\fnm{Wei} \sur{Liu}}\email{wei.liu@uts.edu.au}
\affil[1]{\orgdiv{School of Computer Science}, \orgname{University of Technology Sydney}, \orgaddress{\street{15 Broadway}, \city{Sydney}, \postcode{2007}, \state{NSW}, \country{Australia}}}
\affil[2]{\orgdiv{School of Software}, \orgname{Shandong University}, \orgaddress{\street{27 Shanda Nanlu}, \city{Jinan}, \postcode{250100}, \state{Shandong}, \country{China}}}
% \author*[1,2]{\fnm{First} \sur{Author}}\email{iauthor@gmail.com}



% \affil*[1]{\orgdiv{Department}, \orgname{Organization}, \orgaddress{\street{Street}, \city{City}, \postcode{100190}, \state{State}, \country{Country}}}

% \affil[2]{\orgdiv{Department}, \orgname{Organization}, \orgaddress{\street{Street}, \city{City}, \postcode{10587}, \state{State}, \country{Country}}}

% \affil[3]{\orgdiv{Department}, \orgname{Organization}, \orgaddress{\street{Street}, \city{City}, \postcode{610101}, \state{State}, \country{Country}}}


\abstract{Deep learning underpins most of the currently advanced natural language processing (NLP) tasks such as textual classification, neural machine translation (NMT), abstractive summarization and question-answering (QA). However, the robustness of the models, particularly QA models, against adversarial attacks is a critical concern that remains insufficiently explored. This paper introduces QA-Attack (Question Answering Attack), a novel word-level adversarial strategy that fools QA models. Our attention-based attack exploits the customized attention mechanism and deletion ranking strategy to identify and target specific words within contextual passages. It creates deceptive inputs by carefully choosing and substituting synonyms, preserving grammatical integrity while misleading the model to produce incorrect responses. Our approach demonstrates versatility across various question types, particularly when dealing with extensive long textual inputs. Extensive experiments on multiple benchmark datasets demonstrate that QA-Attack successfully deceives baseline QA models and surpasses existing adversarial techniques regarding success rate, semantics changes, BLEU score, fluency and grammar error rate.}




\maketitle
\section{Introduction}\label{sec1}
% In recent years, adversarial attacks have shown vulnerabilities in natural language processing (NLP) models by slightly modifying the input \cite{modernqa_survey}. Adversarial examples for text classification \cite{bert_attack}, machine translation \cite{Seq2Sick,attack-translation}, and sentiment analysis\cite{tan-etal-2020-morphin} models compromise the performance of textual NLP systems. The increasing vulnerability poses substantial weakness for NLP models. \cite{qa_review}. Question Answering (QA) is one of the vital tasks in Sequence-to-Sequence (Seq2Seq) models with increasing demand in the real world, such as customer service chatbots, search engines, and medical or legal information retrieval systems \cite{qa_usage}. However, if these systems are maliciously attacked, it could lead to misinformation, privacy breaches, or compromised decision-making in critical domains. Consequently, in this paper, we focus on conducting adversarial attacks on question-answering models.

Question-answering (QA) models, a key task within Sequence-to-Sequence (Seq2Seq) frameworks, aim to enhance computers’ ability to process and respond to natural language queries. As these models have evolved, they have been widely adopted in real-world applications such as customer service chatbots\cite{chatbot}, search engines \cite{search_engine}, and information retrieval in fields like medicine \cite{medicine} and law \cite{legal}. However, despite the significant progress in deep learning and natural language processing (NLP), these models remain vulnerable to adversarial examples, leading to misinformation, privacy breaches, and flawed decision-making in critical areas \cite{qa_usage, dong2022adversarial, hathaliya2022adversarial, sun2021adversarial}. This highlights the importance of understanding how adversarial examples are generated from the attackers’ perspective and potential defense mechanisms — an area that remains under-explored.

% Question Answering (QA) models are designed to comprehend given texts and questions to provide accurate and reasonable answers \cite{qa_review}. Current QA models primarily focus on two types of questions: \textbf{Informative Queries} and \textbf{Boolean Queries}. Informative Queries typically begin with interrogative words such as ``who'', ``what'', ``where'', ``when'', ``why'', or ``how'' and seek detailed information from the given context. In contrast, Boolean Queries require a binary response of either ``Yes'' or ``No''. Current state-of-the-art question-answering baselines include encoder-decoder models \cite{T5, LongT5, BART}, BERT models\cite{bert, RoBERTa}, and some GPT models \cite{gpt}.

QA models are expected to comprehend given texts and questions, providing accurate and contextually relevant answers \cite{qa_review}. These models primarily address two types of questions: {Informative Queries} and {Boolean Queries}. The {Informative Queries} typically begin with interrogative words such as “who,” “what,” “where,” “when,” “why,” or “how,” requiring detailed and specific information from the provided context. Although models like T5 \cite{T5}, LongT5 \cite{LongT5}, and BART \cite{BART}, which follow an encoder-decoder structure, have demonstrated strong performance, they still suffer from the maliciously crafted adversarial examples. Initially, studies like ``Trick Me If You Can'' \cite{wallace2019trick} primarily relied on human annotators to construct effective adversarial question-answering examples. This methodology, however, inherently constrained scalability and increased resource demands. As research progressed, automated approaches for attacking textual classifiers in QA models emerged. Gradient-based methods, as employed in RobustQA \cite{robustmultilingual}, UAT \cite{GBWI}, and HotFlip \cite{hotflip}, were developed to identify and modify the most influential words affecting model answers. Building upon a deeper understanding of QA tasks, subsequent studies explored more targeted strategies. For instance, Position Bias \cite{ko-etal-2020-look}, TASA \cite{TASA}, and Entropy Maximization \cite{entroph-max} investigated the manipulation of sentence locations and the analysis of answer sentences to identify vulnerable parts of the context. These approaches refined the attack methods by applying modifications through paraphrasing or replacing original sentences, thus enhancing the effectiveness of adversarial examples. However, these methods encounter two primary challenges: 1) None of these attack methods is suitable for both ``informative queries'' and ``boolean queries''. 2) Constraining the search space for optimal vulnerable words to answer-related sentences compromises attack effectiveness; meanwhile, targeting entire sentences proves inefficient \cite{jia-liang-2017-adversarial}.

In addition, {Boolean Queries} seek a simple binary ``Yes'' or ``No'' answer. Models like BERT \cite{bert}, RoBERTa \cite{RoBERTa}, and GPT variants \cite{gpt2, gpt3, gpt4, chatgpt}, which excel at sentence-level understanding and token classification, are widely used for Boolean QA tasks. These models leverage their deep contextual understanding of language to accurately determine whether a given statement is true or false, making them state-of-the-art baselines for the task.
Researchers have proposed various approaches to target boolean classifiers in the context of Boolean Queries attacks. Attacks like \cite{bert_attack, bae, cla_att_random, pso, pwws}, which involve adding, relocating, or replacing words, are based on the influence that each word has on the prediction. They retrieve word importance by the output confidence to the level or with gradient. However, gradient calculation is computationally intensive and ineffective when dealing with long context input, and knowing victim models' internal information is unrealistic in practice.

We present QA-Attack, an adversarial attack framework tailored for both {Informative Queries} and {Boolean Queries} in QA models. QA-Attack uses a Hybrid Ranking Fusion (HRF) algorithm that integrates two methods: Attention-based Ranking (ABR) and Removal-based Ranking (RBR). ABR identifies important words by analyzing the attention weights during question processing, while RBR evaluates word significance by observing changes in the model’s output when specific words are removed. The HRF algorithm combines these insights to locate vulnerable tokens, which are replaced with carefully selected synonyms to generate adversarial examples. These examples mislead the QA system while preserving the input's meaning. This unified attack method improves both performance and stealth, ensuring realistic applicability for both types of queries.
In summary, our work makes the following key contributions:
\begin{itemize}
    \item We present QA-Attack with a Hybrid Ranking Fusion (HRF) algorithm designed to target question-answering models. This novel approach integrates attention and removal ranking techniques, accurately locating vulnerable words and fooling the QA model with a high success rate.
    \item Our QA-Attack can effectively target multiple types of questions. This adaptability allows our method to exploit vulnerabilities across diverse question formats, which significantly broadens the scope of potential attacks in various real-world scenarios.
    % \item Contribute to the HRF algorithm, QA-Attack has stunning performance on attacking long context.\luk{remove}
    \item QA-Attack generates adversarial examples by implementing subtle word-level changes that preserve both linguistic and semantic integrity while minimizing the extent of alterations, and we conduct extensive experiments on multiple datasets and victim models to thoroughly evaluate our method's effectiveness in attacking QA models.
\end{itemize}

The rest of this paper is structured as follows. We first review QA system baselines and adversarial attacks for QA models in Section \ref{Related Work}. Then we detail our proposed method in Section \ref{Our Proposed Attack Method}. We evaluate the performance of the proposed method through extensive empirical analysis in Section \ref{Experiment and Analysis}. We conclude the paper with suggestions for future work in Section \ref{Conclusion and Future Work}.
\section{Related Work}\label{Related Work}
This section provides a comprehensive overview of question-answering models and examines the existing research on adversarial attacks against them.
\subsection{Question Answering Models}
Question answering represents a complex interplay of NLP, information retrieval, and reasoning capabilities \cite{qa_review,qa_survey}. Basically, these models are designed to process an input question and a context passage, extracting or generating an appropriate answer through elaborate analysis of the semantic relationships between these elements \cite{modernqa_survey}. Modern QA systems typically rely on deep learning models with transformer-based architectures like BERT \cite{bert} and its variants \cite{DistilBERT, RoBERTa, ALBERT} being particularly prevalent. These models excel at capturing contextual information and understanding nuanced relationships in the text with transformers, allowing them to perform impressively on QA tasks. In addition to these transformer models, encoder-decoder architectures such as T5 \cite{T5, UnifiedQA} and BART \cite{BART}, GPT \cite{gpt} and PEGASUS \cite{pegasus} have also become prominent in QA models. These models utilize an encoder to process the input question and context, transforming them into a rich, context-aware representation, and the decoder is then used to generate a coherent and contextually appropriate answer.

\subsection{Previous Works on Attacking QA Models}
With the development of NLP techniques, recent research has increasingly focused on developing sophisticated textual adversarial examples for QA systems \cite{wallace2019trick}. The inherent differences between ``informative queries'' and ``boolean queries'' necessitate distinct attacking diversities due to their unique answer structures \cite{GBWI}. 
Attacks on boolean QA pairs closely resemble methods used to mislead textual classifiers. These attacks primarily operate at the word level, aiming to manipulate the model's binary (yes/no) output \cite{bert_attack, bae}. 
In contrast, informative queries present a more complex challenge. These attacks frequently target the sentence level, requiring an approach to disrupt the model's comprehensive understanding \cite{attack_nlp}. 

\subsubsection{Boolean Queries Attacks}
Boolean queries are similar to classification tasks in NLP, while the answer is based on two-way input: question and context. They are vulnerable to attacks designed for NLP classifiers when question and context are simply encoded and concatenated. Approaches such as \cite{bert_attack}, \cite{bae}, \cite{cla_att_random}, \cite{pso}, and \cite{pwws} concentrate on altering individual words based on their influence on model predictions. These methods typically employ carefully selected synonyms for word substitution. The process of word replacement is guided either by the direct use of BERT Masked Language Model (MLM) \cite{bert} or by leveraging gradient information to determine optimal substitution candidates.
While effectively fool classifiers (boolean queries), these attacks were initially designed for classification tasks and have shown limited efficacy when applied to the question-and-context format of QA systems. To address this limitation, some attack methods for Seq2Seq models have been adapted for QA models. UAT \cite{GBWI}, which averages gradients and modifies input data to maximize the model's loss, has been adapted for QA but still struggles with boolean queries due to their simplicity. Similarly, TextBugger \cite{TextBugger}, which focuses on character-level perturbations, also faces challenges in handling the deeper semantic understanding required in QA, especially for multi-sentence reasoning. Liang's approach \cite{dtccf}, relying on confidence-based manipulations, has difficulty reducing the model's certainty in boolean queries where the binary answers leave less room for variation in confidence. Although these approaches offer improved accuracy in attacking informative questions with minor modifications, they struggle with boolean queries.
We argue that these methods face challenges in identifying the most vulnerable words when dealing with concatenated question-context input relationships. The MLQA attack \cite{multilingualbert} attempts to bridge this gap by utilizing attention weights to identify and alter influential words. However, this method, developed specifically for multi-language BERT models, may not fully address QA-specific vulnerabilities.

\subsubsection{Informative Queries Attacks}
In contrast to boolean queries, adversarial attacks on informative queries within QA systems share fundamental similarities with attacks on other Seq2Seq models \cite{summarization_attack, att-translate, attention_translation}, concentrating more on the inter-relationship between question and context. The defense mechanisms like RobustQA \cite{robustmultilingual} have been developed to enhance model resilience through improved training methods, and sophisticated attacks continue to successfully compromise these systems, especially when employing subtle manipulations of key input elements.
Character-level attack methods, notably HotFlip \cite{hotflip}, have demonstrated significant success by strategically flipping critical characters based on gradient information, leading to misinterpreting informative inputs. In the multilingual domain, MLQA \cite{MultiQA} leverages attention weights to identify and target crucial words, though its attention mechanism, primarily designed for multilingual functionality, may not fully exploit the intricate vulnerabilities within the model's attention architecture.
Advanced techniques have emerged to target the influence that answers have on QA systems. Position Bias and Entropy Maximization methods exploit model weaknesses by manipulating contextual patterns and answer positioning, particularly effective in scenarios involving complex, lengthy responses. Syntactically Controlled Paraphrase Networks (SCPNs) \cite{iyyer-etal-2018-adversarial} generate adversarial examples through strategic syntactic alterations while preserving semantic meaning. TASA (Targeted Adversarial Sentence Analysis) \cite{TASA} primarily relies on manipulating the answer sentences to mislead QA models, making it particularly effective for informative queries where complex responses provide more opportunities for subtle modifications. However, this approach is not suitable for boolean queries, as the simplicity of yes/no answers limits the sentence-level manipulations that TASA depends on.

% Despite these advancements, current attack methods face several challenges. They often lack versatility across different query types, struggle with efficiently targeting vulnerable parts in long contexts, and can be computationally intensive, especially for gradient-based approaches. Word-level attacks may struggle to maintain semantic consistency, while sentence-level attacks can introduce more noticeable modifications to the text.

% \subsection{Attention Mechanism}
\section{Our Proposed Attack Method}\label{Our Proposed Attack Method}

\begin{figure}[!t]
    \centering
    \includegraphics[width=0.98\linewidth]{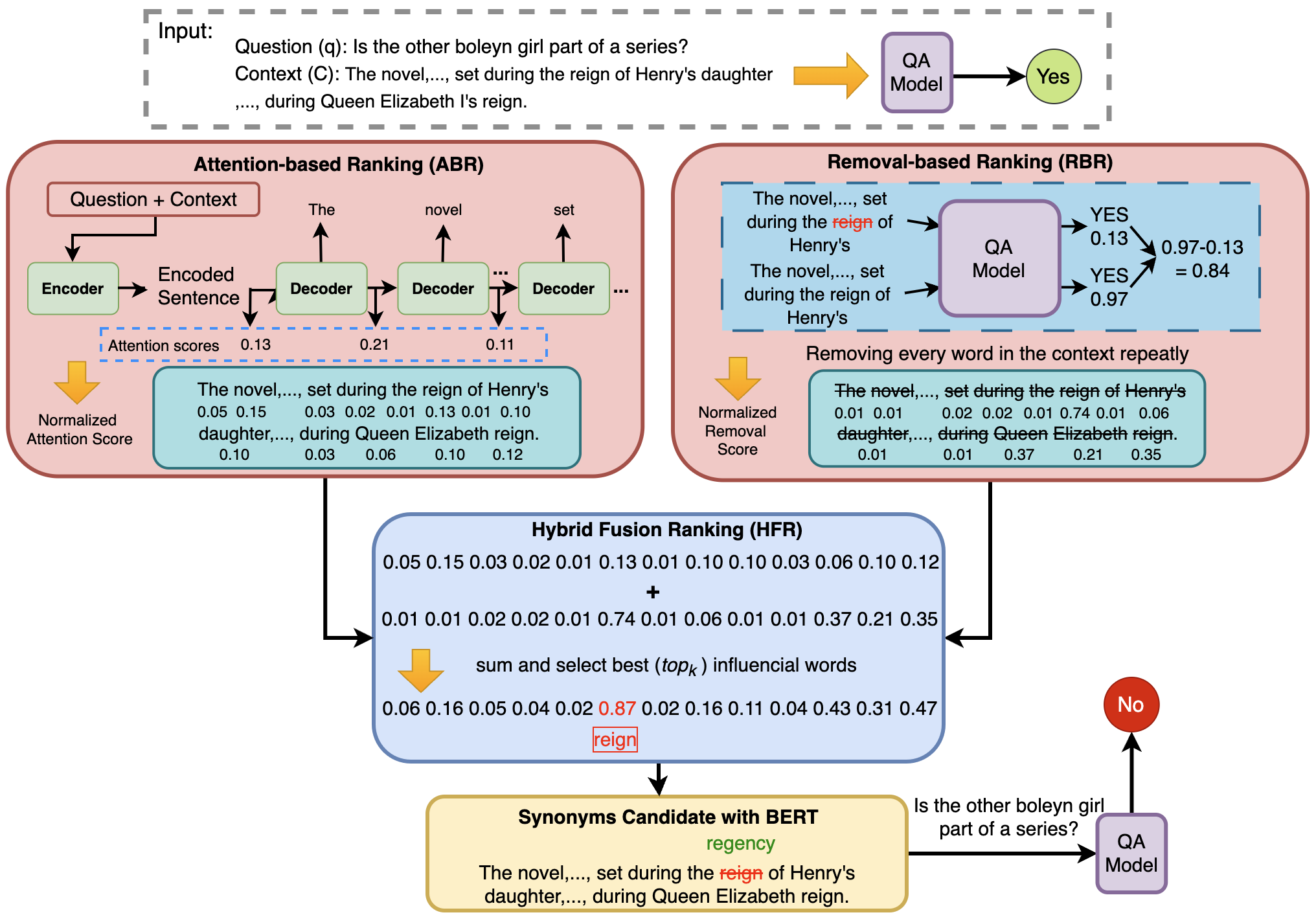}
    \caption{The workflow of our QA-Attack algorithm for QA models. It processes question-context pairs through two parallel modules: Attention-based Ranking (ABR) and Removal-based Ranking (RBR). These modules generate attention and removal scores respectively for each word using customized attention mechanisms and removal ranking strategies. The scores are then aggregated, and the $top_{k}$ highest-scoring words are selected as candidates. Finally, these candidates are replaced with BERT-generated synonyms to create adversarial examples that can effectively mislead the QA model.}
    \label{fig:qa-attack workflow}
\end{figure}

In this section, we introduce the QA-Attack algorithm. It can be summarized into three main steps. First, the method effectively captures important words in context by processing pairs of questions and corresponding context using attention-based and removal-based ranking approaches. Then, attention and removal scores are combined, allowing the identification of the most influential words. At last, a masked language model \cite{bert} is utilized to identify potential synonyms that could replace the targeted words. The overall workflow of QA-Attack is shown in Figure \ref{fig:qa-attack workflow}. In the following sections, we explain our model in detail.

\subsection{Problem Setting}
\RestyleAlgo{ruled}
\begin{algorithm}[!t]
    \caption{QA-Attack Algorithm}
    \label{alg: QA-Attack algorithm}
    \SetKwInOut{Input}{Input}\SetKwInOut{Output}{Output}
    \Input{QA victim model $F(\cdot)$, logits $L$, question $q$, context $C$,  words in the context $c$, reference answer $a$, attention network $A$, top $k$ words to attack $top_k$, number of synonyms $d$, BERT MLM model $BERT$ for generating synonyms.}
    \Output{Optimal adversarial sample $C'$}
    
    \textbf{// Attention-based Ranking}\;
    Compute attention scores: $\alpha \gets [(c, A(q + C))]$\;
    Initialize attention score list: $attention\_scores \gets [\ ]$\;
    
    \For{each $score$ in $\alpha$}{
    \If{$score\in C$}{
        Append $score$ to $attention\_scores$\;
        }
    }
    
    \textbf{// Removal-based Ranking}\;
    Initialize importance score list: $importance\_scores \gets [\ ]$\;
    
    \For{each $c$ in $C$}{
        Generate modified context: $C^* \gets C$ excluding $c$\;
        Compute importance score: $importance\_scores.append(|F(q, C^*) - F(q, C)|)$\;
    }
    
    \textbf{// Hybrid Ranking Fusion}\;
    Combine attention and importance scores: $combined\_scores \gets attention\_scores \cup importance\_scores$\;
    Select $top_k$ words: $top\_k\_list \gets \text{sort}(combined\_scores)[\ :top_k]$\;
    
    Initialize adversarial examples list: $Adv\_list \gets [\ ]$\;
    
    \For{each $t$ in $top\_k\_list$}{ 
       Generate adversarial token from $d$ potential synonyms: $c_{adv} \gets BERT(t)$\; 
       Create adversarial context: $\Delta C \gets [c_1, \ldots, c_{adv}, \ldots, c_n]$\;
       Append $\Delta C$ to $Adv\_list$\;
    }
    Initialize maximum gap: $\text{max\_gap} \gets -\infty$\;
    Initialize optimal adversarial context: $C' \gets \emptyset$\;
    \For{each $adv$ in $Adv\_list$}{
        \If{$F(adv) \neq a$}{
            Compute gap: $\text{gap} \gets L(F(adv)) - L(F(C))$\;
            \If{$\text{gap} > \text{max\_gap}$}{
                Update maximum gap: $\text{max\_gap} \gets \text{gap}$\;
                Update optimal adversarial context: $C' \gets adv$\;
            }
        }
    }
    \Return Optimal adversarial sample $C'$
\end{algorithm}

Given a pre-trained question-answering model $F$, which receives an input of context $C$, question $q$, and outputs answer $a$, such that \( F(q, C) = a\). The objective is to deceive the performance of $F$ with perturbed context $C'$ such that \( F(q, C') \neq a\). To craft $C'$, a certain number of perturbation \( c_{adv} \) is added to the context $C$ by replacing some of its original tokens \{$c_1, c_2, ..., c_n$\}.

\subsection{Attention-based Ranking (ABR)}\label{Attention-based Ranking}
% \begin{figure}[!t]
%     \centering
%     \includegraphics[width=0.95\linewidth]{images/attention_scores.png}
%     \caption{An example of word-level attention scores match on both context and question. All score values are normalized to 0 to 1. The blue sentence is the question, and the green sentence is the context.}
%     \label{fig: Attention scores}
% \end{figure}
Attention mechanisms were first used in image feature extraction in the computer vision field \cite{att_road_sign,show_att_tell,attention_review}. However, they were later employed by \cite{att-translate} to solve machine translation problems. In translation tasks, attention mechanisms enable models to prioritize and focus on the most relevant parts of the input data \cite{attention_translation}. In question-answering tasks, attention scores are imported to examine the relationships between question and context, allowing the model to determine which words or phrases are most relevant to answering the question \cite{transformers}. Hence, we leverage the attention score to identify target words for our attack. We employ the attention mechanism from T5 \cite{T5} that has been specifically optimized for question-answering tasks in UnifiedQA \cite{UnifiedQA}.
% QA-Attack applied the attention network used in the T5 model\cite{T5}. 
As shown in Fig. \ref{fig:qa-attack workflow}, the ``Attention-based Ranking'' begins by encoding the input context and question through an encoder. During the encoding process, self-attention allows the model to analyze how each word in the input relates to every other word, effectively highlighting the words that carry the most weight in understanding both question and context. In the decoding process, cross-attention further refines this by focusing on the parts of the input most relevant to generating the correct output. By averaging the attention scores of all layers and heads, we match them to each input word.
% This encoded sentence is then passed to a decoder, where each word receives an attention score based on its importance in generating the answer. The architecture allows for comprehensive information processing: in the encoder, each position can focus on all positions from the previous encoder layer, while in the decoder, each position can attend to all positions up to and including its own position in the previous decoder layer.

The implement details are shown in Algorithm~\ref{alg: QA-Attack algorithm}. The question \& context pair is fitted into attention network $A$, and we filter out the attention scores for context (lines 1 to 8 of Algorithm~\ref{alg: QA-Attack algorithm}). Then, the attention score of each word corresponding to each layer is summed up. After averaging and normalization, the word-level attention score is obtained.
% , as shown in Figure \ref{fig: Attention scores}.

\subsection{Removal-based Ranking (RBR)}\label{Removal-based Ranking}
Previous studies on adversarial attacks in the text have shown that each word's significance can be quantified using an importance score \cite{cla_att_random, summarization_attack, TASA, bert_attack}. This score is largely determined by how directly the word influences the final answer. To enhance the efficacy of ranking progress, we rank each word in the context to obtain the removal importance score (lines 9 to 14 of Algorithm~\ref{alg: QA-Attack algorithm}). Given the input context $C$ containing $n$ words from $c_1$ to $c_n$ and question $q$, the importance score (removal score) of the $i \ th$ ($1 \leq i \leq n$) word $c_i$ is:
\begin{equation}
\label{equation: removal importance score}
%\begin{split}
I_i = L_{F}(a \mid q,C) - L_{F}(a \mid q, C \setminus c_i), \\
%\end{split}
\end{equation}
where $C \setminus c_i$  represents the context after deleting $c_i$, and $L_{F} = \log P(a \mid q, C)$ refers to the probability (logits) of the label, respectively.
\subsection{Hybrid Ranking Fusion (HRF)}
The attention-based and removal-based word selection techniques offer complementary perspectives on token significance, each highlighting different aspects of word importance. Consequently, we tend to choose words that both methods consider significant. This is achieved by adding the scores from each method for every word to create a fusion score.

When generating a fusion score, we address several key factors. First, we independently normalize the attention and removal scores before adding them together. Then, to balance attack effectiveness and efficiency, we introduce a $top_k$ parameter, a positive integer that controls the number of words targeted. Finally, we select the $top_k$ highest-scoring words for modification (lines 15 to 18 of Algorithm \ref{alg: QA-Attack algorithm}).

\subsection{Synonym Selection}
Various synonym generation methods exist, including Word2Vec \cite{word2vec}, Hownet \cite{hownet}, and WordNet \cite{hownet}. We adopt BERT \cite{bert} for synonym selection due to its textual capabilities, which enable it to generate synonyms based on the complete sentence structure. Unlike Word2Vec's static embeddings or WordNet's fixed synonym lists, BERT's context-sensitive approach allows for dynamic synonym selection that preserves both semantic meaning and grammatical correctness. This contextual awareness makes BERT particularly effective for crafting natural and semantically coherent adversarial examples.

We process each selected word in the context by replacing it with the ``[MASK]'' token. This modified context is then input into the BERT Masked Language Model (MLM) to predict the most likely substitutions for the masked word. To expand the range of potential samples, we introduce a parameter $d$ that controls the number of synonym substitutions considered (lines 19 to 23 of Algorithm \ref{alg: QA-Attack algorithm}). 
% For each of the $top_k$ targeted words, $d$ synonym candidates are generated, resulting in $k \times d$ potential adversarial samples per input 
This approach allows us to generate a diverse set of imperceptible replacements while maintaining contextual relevance.
\subsection{Candidate Selection}
We define an optimal adversary as one that maximizes the difference between the predicted answer and the attacked answer. For boolean queries, following textual classifier approaches that utilize logits to decide output label (yes/no), we compare the logits of output answers. For informative queries, we sum the logits of individual words. Using $L$ to denote the logits derivation function, we identify the optimal adversary from the ``Adv\_list'' as shown in lines 24 to 35 of Algorithm~\ref{alg: QA-Attack algorithm}.

\section{Experiment and Analysis}\label{Experiment and Analysis}
In this section, we present a comprehensive evaluation of QA-Attack's performance compared to current state-of-the-art baselines. Our analysis covers several key aspects with various metrics, providing a thorough understanding of our method's capabilities, limitations, and performance across diverse scenarios. We provide a detailed analysis of attack performance and imperceptibility (Sec. \ref{experiment analysis}). Besides, to gain deeper insights, we conduct ablation studies (Sec. \ref{Ablation Studies}) and assess attacking efficiency (Sec. \ref{efficiency}). In addition, we examine QA-Attack's response to defense strategies (Sec. \ref{defense}), exploring the effects of adversarial retraining (Sec. \ref{Adversarial Retraining}) and investigating the transferability of attacks (Sec. \ref{Transferability of Attacks}). Additionally, we report the preference of our attack by investigating parts of speech preference (Sec. \ref{Parts of speech preference}) and analyzing its robustness versus the scale of pre-trained models (Sec. \ref{Robustness versus the scale of pre‑trained models}). 
% This comprehensive evaluation not only demonstrates QA-Attack's effectiveness but also offers valuable insights into its adaptability and performance in various contexts within the landscape of adversarial attacks on question-answering systems.

\subsection{Experiment Settings and Evaluation Metrics}\label{Experiment Settings and Evaluation Metrics}
The base setting of our experiments is let $top_k = 5$, $d = 2$, and use a BERT-base-uncased\footnote{\url{https://github.com/google-research/bert/?tab=readme-ov-file}.} with 12 Transformer encoder layer (L) and 768 hidden layers (H) as the synonym generation model. Tables \ref{UnifiedQA_results}, \ref{Bert_results}, and \ref{longt5_results} summarize the experimental results on {informative queries} datasets, offering a comparative analysis of our QA-Attack method against five state-of-the-art QA baselines. For {boolean queries}, we present the attacking results on the BoolQ dataset in Table \ref{BoolQ_results}. Some visualised examples are shown in Table \ref{tab: visualized results}. Besides, we provide code for the reproductivity of our experiments\footnote{Our code is available at: \url{https://github.com/UTSJiyaoLi/QA-Attack}.}. The metrics used in our experiment are:
\begin{itemize}
    \item \textbf{F1}: The F1 score balances precision and recall, providing a nuanced view of how much the attacked answers match reference answers.
    \item \textbf{ROUGE and BLEU}: A higher BLEU \cite{bleu} or ROUGE \cite{rouge} score in context
    % , reported in Table \ref{UnifiedQA_results}, \ref{Bert_results}, and \ref{longt5_results}, 
    indicates that the adversarial context retains more of the exact phrasing, contributing to better linguistic fluency and coherence. 
    % In contrast, ,lower ROUGE and BLEU scores, reported in Table \ref{} for answers indicate better attack performance.
    \item \textbf{Exact Match (EM)} Measures the percentage of model predictions that exactly match the correct answers in both content and format.
    \item \textbf{Similarity (SIM)}:  Evaluates the semantic similarity between original and adversarial context using BERT \cite{bert} embeddings. (Note: In our following experiments, EM and SIM are not only measured answers but also reflect the quality of the generated context in Sec. \ref{Texual quality of the word candidates}).
    \item \textbf{Modification Rate (Mod)}:  Mod measures the proportion of altered tokens in the text. This metric considers each instance of replacement, insertion, or deletion as a single token modification.
    \item \textbf{Grammar Error (GErr)}: GErr measures the increase in grammatical inaccuracies within successful adversarial examples relative to the original text. This measurement employs LanguageTool \cite{grammar} to enumerate grammatical errors.
    \item \textbf{Perplexity (PPL)}: PPL serves as an indicator of linguistic fluency in adversarial examples \cite{kann-etal-2018-sentence,pso}. The perplexity calculation utilizes a GPT-2 model with a restricted vocabulary \cite{radford2019language}.
\end{itemize}

\subsection{Datasets and Victim Models}
\begin{table}[!t]
\centering
\caption{Dataset distribution and corresponding baseline performance (F1).}
\label{table:qa_datasets}
\begin{tabular}{cccccccc}
    \toprule
    \multirow{2}*{Dataset}&\multicolumn{4}{c}{Data Distribution}&\multicolumn{3}{c}{Model Performance (F1)}\\
    \cmidrule(lr){2-5} \cmidrule(lr){6-8}
     & Total & Train & Validation & Test & T5 & LongT5 & BERT\textsubscript{base}\\ 
    \midrule
    SQuAD 1.1 & ~100,000 & ~87,600 & ~10,570 & N/A & 88.9 & 89.5 & 88.5 \\
    SQuAD V2.0 & ~150,000 & ~130,319 & ~11,873 & N/A & 81.3 & 83.2 & 74.8 \\
    NewsQA & ~119,000 & ~92,549 & ~5,165 & ~5,126 & 66.8 & 67.2 & 60.1 \\
    BoolQ & ~16,000 & ~9,427 & ~3,270 & ~3,245 & 85.2 & 86.1 & 80.4 \\
    NarrativeQA & ~45,000 & ~32,747 & ~3,461 & ~10,557 & 67.5 & 68.9 & 62.1 \\
    \bottomrule
\end{tabular}
\end{table}

We assess QA-Attack using four {informative queries} datasets: SQuAD 1.1 \cite{SQuAD1.1}, SQuAD V2.0 \cite{SQuADv2}, NarrativeQA \cite{NarrativeQA}, and NewsQA \cite{NewsQA}, along with the {boolean queries} dataset BoolQ \cite{BoolQ}.
\begin{itemize}
    \item SQuAD 1.1: Questions formulated by crowd workers based on Wikipedia articles. Answers are extracted as continuous text spans from the corresponding passages.
    \item SQuAD 2.0: Extension of SQuAD 1.1 incorporating unanswerable questions. These questions are designed such that no valid answer can be located within the provided passage, adding complexity to the task.
    \item NarrativeQA: Questions based on entire books or movie scripts. Answers are typically short and abstractive, demanding deeper comprehension and synthesis of narrative elements.
    \item NewsQA: Questions based on CNN news articles designed to test reading comprehension in the context of current events and journalistic writing.
    \item BoolQ: Dataset of boolean (yes/no) questions derived from anonymized, aggregated queries submitted to the Google search engine, reflecting real-world information-seeking behaviour.
\end{itemize}
Our experiment includes three question-answering models for comparison. They are T5\cite{UnifiedQA}, LongT5 \cite{LongT5}, and BERT\textsubscript{base} \cite{bert}. The LongT5 is an extension of T5 with an encoder-decoder specifically for long contextual inputs. The BERT-based models are structured with bidirectional attention, meaning each word in the input sequence contributes to and receives context from both its left and right sides. Table \ref{table:qa_datasets} presents the distribution of dataset splits and F1 scores reported on each QA baseline.

\subsection{Baseline Attacks}

For our experimental baselines, we employ five leading attack methods: TASA \cite{TASA}, RobustQA \cite{robustmultilingual}, Tick Me If You Can (TMYC) \cite{wallace2019trick}, T3 \cite{t3}, and TextFooler \cite{cla_att_random}. We utilize the official implementation of T3 in its black-box setting, while TASA, TMYC, and RobustQA are employed with their standard configurations. TextFooler, originally not designed for question-answering tasks, was adapted for our experiments. We modified it to process the context only (questions are removed).

% \begin{equation}
% \label{BLEU and ROUGE}
% \begin{split}
% \text{BLEU} &= \frac{\sum_{C \in \{Candidates\}} \sum_{w \in C} \text{Count}_{\text{clip}}(w)}{\sum_{C \in \{Candidates_\}} \sum_{w \in C} \text{Count}(w)}\\
% \text{ROUGE}&=\frac{\sum_{\substack{S \in \text{ReferenceAnswers}}} \sum_{\substack{\text{uni-gram} \in S}} \text{Count}_{\text{match}}(\text{uni-gram})}{\sum_{\substack{S \in \text{ReferenceAnswer}}} \sum_{\substack{\text{uni-gram} \in S}} \text{Count}(\text{uni-gram})}
% % \text{BLEU-2} &= \left(\text{Precision}_{\text{unigram}} \times \text{Precision}_{\text{bigram}}\right)^{\frac{1}{2}}
% \end{split}
% \end{equation}

\subsection{Experiment Analysis}\label{experiment analysis}

\begin{table}[!ht]
    \centering
    \caption{Comparison of original and adversarial contexts for two types of queries. The table highlights the differences between the original and adversarial contexts, as well as the corresponding answers provided by the model before and after the attack.}
%\small
\begin{tabularx}{\textwidth}{l X}
    \toprule
        \textbf{Question} & Was the movie ``The Strangers'' based on a true story? \\
        \hdashline
        \textbf{Context} & The Strangers is a 2008 American slasher \textcolor{blue}{film} written and directed by Bryan Bertino. Kristen (Liv Tyler) and James (Scott Speedman) are \textcolor{blue}{expecting} a relaxing weekend at a family vacation home, but their stay turns out to be anything but peaceful as three masked torturers leave Kristen and James struggling for survival. Writer-director Bertino was \textcolor{blue}{inspired} by real-life \textcolor{blue}{events}: the Manson family Tate murders, a multiple homicide; the Keddie Cabin Murders, that occurred in California in 1981; and a series of break-ins that occurred in his own \textcolor{blue}{neighborhood} as a child.  \\
        \hdashline
        \textbf{Adversary} & The Strangers is a 2008 American slasher \textcolor{red}{thriller} written and directed by Bryan Bertino. Kristen (Liv Tyler) and James (Scott Speedman) are \textcolor{red}{spending} a relaxing weekend at a family vacation home, but their stay turns out to be anything but peaceful as three masked torturers leave Kristen and James struggling for survival. Writer-director Bertino was \textcolor{red}{influenced} by real-life \textcolor{red}{incidents}: the Manson family Tate murders, a multiple homicide; the Keddie Cabin Murders, that occurred in California in 1981; and a series of break-ins that occurred in his own \textcolor{red}{home} as a child.  \\
        \hdashline
        \textbf{Original Answer} & Yes \\
        \hdashline
        \textbf{Attacked Answer} & No \\
        \bottomrule
        \toprule
        \textbf{Question} & Who ruled the Duchy of Normandy? \\
        \hdashline
        \textbf{Context} & The Normans were famed for their \textcolor{blue}{martial} spirit ... The Duchy of Normandy, which they formed by treaty with the French crown, was a great fief of medieval France, and under \textcolor{blue}{Richard} I of Normandy was forged into a cohesive and formidable principality in feudal tenure ... Norman adventurers \textcolor{blue}{founded} the Kingdom of Sicily ... an \textcolor{blue}{expedition} on behalf of their duke, William the Conqueror, led to the Norman conquest of England at the \textcolor{blue}{Battle} of Hastings in 1066. \\
        \hdashline
        \textbf{Adversary} & The Normans were famed for their \textcolor{red}{warrior} spirit ... The Duchy of Normandy, which they formed by treaty with the French crown, was a great fief of medieval France, and under \textcolor{red}{William} I of Normandy was forged into a cohesive and formidable principality in feudal tenure ... Norman adventurers \textcolor{red}{invaded} the Kingdom of Sicily ... an \textcolor{red}{invasion} on behalf of their duke, William the Conqueror, led to the Norman conquest of England at the \textcolor{red}{siege} of Hastings in 1066. \\
        \hdashline
        \textbf{Original Answer} & The French crown \\
        \hdashline
        \textbf{Predicted Answer} & William I of Normandy \\
        \bottomrule
    \end{tabularx}
    \label{tab: visualized results}
\end{table}

\begin{table}[!ht]
\centering
\caption{Comparative analysis of QA-Attack and baseline models on T5. Drops of BLEU and ROUGE scores (uni-gram) on contexts are reported in the table, with higher values indicating better performance. For F1, EM, and SIM (i.e., similarity) metrics on answers, lower values indicate better performance.}
\label{UnifiedQA_results}
%\small
\begin{tabular}{@{}p{1.5cm}p{4.3cm}ccccc@{}} % Define column alignment
    \toprule
    Datasets & Methods & F1$\downarrow$ & EM$\downarrow$ & ROUGE$\uparrow$ & BLEU$\uparrow$ & SIM$\downarrow$ \\ \midrule
    \multirow{6}{*}{SQuAD 1.1} 
    & TASA~\cite{TASA} & 9.21 & 7.49 & 89.12 & 82.88 & 6.38 \\
    & TMYC (Tick Me If You Can)~\cite{wallace2019trick} & 7.28 & 8.21 & 81.91 & 78.72 & 8.22 \\
    & RobustQA~\cite{robustmultilingual} & 5.89 & 7.52 & 84.23 & 77.41 & 6.03 \\
    & TextFooler~\cite{cla_att_random} & 10.6 & 10.49 & 83.11 & 76.05 & 6.29 \\
    & T3~\cite{t3} & 5.41 & 6.29 & 86.83 & 73.82 & 7.23 \\
    & QA-Attack (ours) & \textbf{4.67} & \textbf{5.68} & \textbf{90.51} & \textbf{84.11} & \textbf{5.91} \\ \midrule

    \multirow{6}{*}{SQuAD V2.0} 
    & TASA~\cite{TASA} & 20.09 & 19.31 & 70.21 & 76.06 & 7.29 \\
    & TMYC (Tick Me If You Can)~\cite{wallace2019trick} & 17.23 & 20.68 & 65.19 & 69.82 & 9.05 \\
    & RobustQA~\cite{robustmultilingual} & 16.37 & 18.73 & 67.71 & 63.19 & 8.14 \\
    & TextFooler~\cite{cla_att_random} & 21.69 & 24.5 & 65.33 & 65.01 & 9.32 \\
    & T3~\cite{t3} & 11.19 & 19.68 & 69.71 & 73.53 & 8.82 \\
    & QA-Attack (ours) & \textbf{9.13} & \textbf{15.41} & \textbf{72.76} & \textbf{77.28} & \textbf{6.33} \\ \midrule

    \multirow{6}{*}{Narrative QA} 
    & TASA~\cite{TASA} & 11.79 & 15.25 & 68.11 & 70.36 & 6.11 \\
    & TMYC (Tick Me If You Can)~\cite{wallace2019trick} & 12.73 & 9.32 & 65.91 & 67.22 & 7.61 \\
    & RobustQA~\cite{robustmultilingual} & 10.01 & 13.91 & 67.19 & 64.11 & 6.81 \\
    & TextFooler~\cite{cla_att_random} & 14.72 & 18.61 & 63.85 & 62.82 & 11.74 \\
    & T3~\cite{t3} & 11.74 & 11.37 & 62.34 & 60.17 & 6.28 \\
    & QA-Attack (ours) & \textbf{5.61} & \textbf{7.23} & \textbf{69.18} & \textbf{75.73} & \textbf{5.23} \\ \midrule

    \multirow{6}{*}{NewsQA}
    & TASA~\cite{TASA} & 8.56 & 29.44 & 77.28 & 69.44 & 7.11 \\
    & TMYC (Tick Me If You Can)~\cite{wallace2019trick} & 6.12 & 31.23 & 77.96 & 72.49 & 9.22 \\
    & RobustQA~\cite{robustmultilingual} & 5.12 & 29.48 & 83.81 & 79.82 & 10.84 \\
    & TextFooler~\cite{cla_att_random} & 9.01 & 30.86 & 74.21 & 57.44 & 27.91 \\
    & T3~\cite{t3} & 6.21 & 28.52 & 75.22 & 72.56 & 14.27 \\
    & QA-Attack (ours) & \textbf{3.61} & \textbf{24.42} & \textbf{78.85} & \textbf{82.83} & \textbf{8.92} \\ 
    \bottomrule
\end{tabular}
\end{table}

\begin{table}[!ht]
\centering
\caption{Comparative analysis of QA-Attack and baseline models on Bert\textsubscript{base}. Drops of BLEU and ROUGE scores (uni-gram) on contexts are reported in the table, with higher values indicating better performance. For F1, EM, and SIM (i.e., similarity) metrics on answers, lower values indicate better performance.}
\label{Bert_results}
%\small
\begin{tabular}{@{}p{1.5cm}p{4.3cm}ccccc@{}} % Define column alignment
    \toprule
    Datasets & Methods & F1$\downarrow$ & EM$\downarrow$ & ROUGE$\uparrow$ & BLEU$\uparrow$ & SIM$\downarrow$ \\ \midrule
    \multirow{6}{*}{SQuAD 1.1} 
    & TASA~\cite{TASA} & 15.27 & 34.33 & 82.87 & 67.22 & 8.19 \\
    & TMYC (Tick Me If You Can)~\cite{wallace2019trick} & 12.89 & 28.63 & 81.51 & 76.39 & 10.24 \\
    & RobustQA~\cite{robustmultilingual} & 15.72 & 25.38 & 79.28 & 73.27 & 15.81 \\
    & TextFooler~\cite{cla_att_random} & 23.04 & 37.28 & 67.28 & 49.49 & 14.11 \\
    & T3~\cite{t3} & 8.79 & 16.11 & 57.19 & 63.81 & 16.92 \\
    & QA-Attack (ours) & \textbf{6.42} & \textbf{13.31} & \textbf{91.22} & \textbf{77.16} & \textbf{7.43} \\ \midrule

    \multirow{6}{*}{SQuAD V2.0} 
    & TASA~\cite{TASA} & 31.22 & 28.9 & 77.06 & 69.05 & 8.22 \\
    & TMYC (Tick Me If You Can)~\cite{wallace2019trick} & 29.38 & 27.77 & 73.81 & 67.23 & 10.34 \\
    & RobustQA~\cite{robustmultilingual} & 27.64 & 31.82 & 75.67 & 71.42 & 11.23 \\
    & TextFooler~\cite{cla_att_random} & 36.8 & 29.49 & 67.14 & 62.67 & 13.28 \\
    & T3~\cite{t3} & 26.16 & 27.47 & 74.94 & 70.14 & 7.24 \\
    & QA-Attack (ours) & \textbf{22.18} & \textbf{21.5} & \textbf{80.12} & \textbf{75.23} & \textbf{4.11} \\ \midrule

    \multirow{6}{*}{Narrative QA} 
    & TASA~\cite{TASA} & 12.11 & 14.51 & 61.15 & 63.04 & 7.32 \\
    & TMYC (Tick Me If You Can)~\cite{wallace2019trick} & 8.41 & 10.23 & 52.89 & 69.82 & 10.72 \\
    & RobustQA~\cite{robustmultilingual} & 7.24 & 9.43 & 63.81 & 67.43 & 9.53 \\
    & TextFooler~\cite{cla_att_random} & 13.74 & 18.79 & 56.11 & 56.82 & 14.21 \\
    & T3~\cite{t3} & 8.49 & 15.35 & 65.48 & 67.09 & 7.83 \\
    & QA-Attack (ours) & \textbf{3.86} & \textbf{9.34} & \textbf{69.44} & \textbf{71.15} & \textbf{5.61} \\ \midrule

    \multirow{6}{*}{NewsQA}
    & TASA~\cite{TASA} & 16.85 & 20.95 & 68.74 & 69.12 & 15.22 \\
    & TMYC (Tick Me If You Can)~\cite{wallace2019trick} & 15.86 & 31.23 & 77.96 & 72.49 & 9.22 \\
    & RobustQA~\cite{robustmultilingual} & 17.72 & 29.48 & 83.81 & 79.82 & 10.84 \\
    & TextFooler~\cite{cla_att_random} & 24.13 & 22.63 & 59.17 & 61.22 & 31.07 \\
    & T3~\cite{t3} & 21.22 & 22.57 & 65.14 & 67.11 & 18.27 \\
    & QA-Attack (ours) & \textbf{14.91} & \textbf{20.20} & \textbf{70.04} & \textbf{74.87} & \textbf{9.22} \\ 
    \bottomrule
\end{tabular}
\end{table}

\begin{table}[!ht]
\centering
\caption{Comparative analysis of QA-Attack and baseline models on LongT5. Drops of BLEU and ROUGE scores (uni-gram) on contexts are reported in the table, with higher values indicating better performance. For F1, EM, and SIM (i.e., similarity) metrics on answers, lower values indicate better performance.}
\label{longt5_results}
\begin{tabular}{@{}p{1.5cm}p{4.3cm}ccccc@{}}
    \toprule
    Datasets & Methods & F1$\downarrow$ & EM$\downarrow$ & ROUGE$\uparrow$ & BLEU$\uparrow$ & SIM$\downarrow$ \\ \midrule
    \multirow{6}{*}{SQuAD 1.1} 
    & TASA~\cite{TASA} & 10.61 & 22.45 & 80.67 & 70.41 & 11.88 \\
    & TMYC (Tick Me If You Can)~\cite{wallace2019trick} & 12.43 & 29.81 & 75.37 & 63.83 & 13.22 \\
    & RobustQA~\cite{robustmultilingual} & 17.22 & 31.11 & 73.11 & 68.29 & 17.64 \\
    & TextFooler~\cite{cla_att_random} & 35.31 & 44.09 & 57.77 & 49.49 & 25.33 \\
    & T3~\cite{t3} & 9.33 & 24.52 & 49.23 & 60.33 & 20.87 \\
    & QA-Attack (ours) & \textbf{7.38} & \textbf{18.78} & \textbf{84.22} & \textbf{72.67} & \textbf{9.67} \\ \midrule

    \multirow{6}{*}{SQuAD V2.0} 
    & TASA~\cite{TASA} & 30.71 & 30.11 & 64.71 & 67.28 & 9.32 \\
    & TMYC (Tick Me If You Can)~\cite{wallace2019trick} & 34.11 & 33.88 & 64.21 & 65.11 & 14.82 \\
    & RobustQA~\cite{robustmultilingual} & 29.01 & 39.59 & 62.91 & 68.22 & 13.09 \\
    & TextFooler~\cite{cla_att_random} & 38.25 & 34.67 & 60.47 & 64.16 & 15.44 \\
    & T3~\cite{t3} & 30.44 & 30.13 & 65.81 & 63.72 & 8.29 \\
    & QA-Attack (ours) & \textbf{27.11} & \textbf{24.73} & \textbf{77.37} & \textbf{70.32} & \textbf{5.29} \\ \midrule

    \multirow{6}{*}{Narrative QA} 
    & TASA~\cite{TASA} & 8.22 & 10.67 & 69.83 & 65.77 & 9.53 \\
    & TMYC (Tick Me If You Can)~\cite{wallace2019trick} & 9.36 & 11.33 & 63.15 & 64.27 & 14.72 \\
    & RobustQA~\cite{robustmultilingual} & 15.83 & 12.03 & 64.28 & 63.12 & 12.77 \\
    & TextFooler~\cite{cla_att_random} & 12.77 & 14.82 & 62.99 & 54.21 & 17.33 \\
    & T3~\cite{t3} & 8.38 & 8.26 & 63.92 & 66.32 & 8.92 \\
    & QA-Attack (ours) & \textbf{4.62} & \textbf{5.33} & \textbf{70.33} & \textbf{68.32} & \textbf{7.44} \\ \midrule

    \multirow{6}{*}{NewsQA}
    & TASA~\cite{TASA} & 16.85 & 24.54 & 64.83 & 66.81 & 14.82 \\
    & TMYC (Tick Me If You Can)~\cite{wallace2019trick} & 19.28 & 29.01 & 62.88 & 68.67 & 11.43 \\
    & RobustQA~\cite{robustmultilingual} & 17.23 & 27.42 & 58.32 & 57.22 & 13.37 \\
    & TextFooler~\cite{cla_att_random} & 27.22 & 26.39 & 53.33 & 53.01 & 25.82 \\
    & T3~\cite{t3} & 17.83 & 25.87 & 63.25 & 65.43 & 19.27 \\
    & QA-Attack (ours) & \textbf{15.32} & \textbf{24.12} & \textbf{68.23} & \textbf{70.55} & \textbf{10.48} \\ 
    \bottomrule
\end{tabular}
\end{table}

\begin{table}[!ht]
\centering
\caption{Attack performance comparison on baseline models using the BoolQ dataset, with top results highlighted in bold. Note that TASA~\cite{TASA} is not applicable to boolean questions.}
\label{BoolQ_results}
\begin{tabular}{@{}p{1.5cm}p{4.3cm}ccccc@{}}
    \toprule
    Victim Models & Methods & F1$\downarrow$ & EM$\downarrow$ & ROUGE$\uparrow$ & BLEU$\uparrow$ & SIM$\downarrow$ \\ \midrule
    \multirow{6}{*}{UnifiedQA} 
    & TASA~\cite{TASA} & --- & --- & --- & --- & --- \\
    & TMYC (Tick Me If You Can)~\cite{wallace2019trick} & 17.43 & 19.36 & 82.09 & 77.23 & 21.83 \\
    & RobustQA~\cite{robustmultilingual} & 14.33 & 18.92 & 79.15 & 80.33 & 13.22 \\
    & TextFooler~\cite{cla_att_random} & 20.11 & 19.07 & 80.91 & 83.25 & 33.82 \\  
    & T3~\cite{t3} & 15.16 & 14.74 & 71.32 & 68.79 & 15.82 \\ 
    & QA-Attack (ours) & \textbf{8.64} & \textbf{13.9} & \textbf{87.31} & \textbf{86.57} & \textbf{11.42} \\ \midrule

    \multirow{6}{*}{Bert\textsubscript{base}} 
    & TASA~\cite{TASA} & --- & --- & --- & --- & --- \\
    & TMYC (Tick Me If You Can)~\cite{wallace2019trick} & 21.35 & 13.28 & 63.21 & 70.57 & 7.34 \\
    & RobustQA~\cite{robustmultilingual} & 24.81 & 9.21 & 69.22 & 76.01 & 6.67 \\
    & TextFooler~\cite{cla_att_random} & 33.02 & 11.57 & 65.11 & 67.81 & 8.17 \\  
    & T3~\cite{t3} & 22.06 & 11.02 & 76.17 & 74.62 & 6.23 \\ 
    & QA-Attack (ours) & \textbf{18.39} & \textbf{6.51} & \textbf{77.21} & \textbf{78.11} & \textbf{4.66} \\ \midrule

    \multirow{6}{*}{LongT5}
    & TASA~\cite{TASA} & --- & --- & --- & --- & --- \\
    & TMYC (Tick Me If You Can)~\cite{wallace2019trick} & 29.77 & 9.82 & 67.04 & 73.22 & 7.43 \\
    & RobustQA~\cite{robustmultilingual} & 24.56 & 8.21 & 70.49 & 71.83 & 9.33 \\
    & TextFooler~\cite{cla_att_random} & 33.02 & 11.57 & 65.11 & 67.81 & 8.17 \\  
    & T3~\cite{t3} & 22.06 & 11.02 & 76.17 & 74.62 & 6.23 \\ 
    & QA-Attack (ours) & \textbf{18.39} & \textbf{6.51} & \textbf{77.21} & \textbf{78.11} & \textbf{4.66} \\ 
    \bottomrule
\end{tabular}
\end{table}

Our experimental results in Table~\ref{UnifiedQA_results},~\ref{Bert_results},~\ref{longt5_results} demonstrate that QA-Attack consistently outperforms baseline methods across all informative datasets. As shown in Table~\ref{BoolQ_results}, our method achieves superior performance on the boolean dataset, surpassing all baseline approaches in degrading victim models' accuracy (note that TASA is designed only for informative queries; it is incompatible with boolean query attacks). For {informative queries}, comparing performance on attacking LongT5 with SQuAD 1.1 and NarrativeQA datasets (representing shortest and longest contexts) in Table~\ref{longt5_results}, we observe that while F1 and EM scores decrease for longer contexts, QA-Attack maintains superiority over baselines. This indicates our approach's robustness and adaptability to varying context lengths, particularly in long text. The improved performance in longer contexts suggests our HRF approach effectively identifies and targets vulnerable tokens. Regarding semantic consistency, QA-Attack achieves lower similarity scores compared to baseline methods, indicating that the answers generated after the attack deviate more in meaning from the ground truth responses.

Additionally, the quality of the generated adversarial samples is evident from the ROUGE and BLEU scores. Our method consistently achieves higher ROUGE and BLEU scores compared to the baselines, which suggests that the adversarial examples generated by QA-Attack are not only effective in terms of altering the model's output but also maintain a high degree of contextual and linguistic coherence. This is largely due to our synonym selection method, which ensures the replacements are contextually appropriate and semantically relevant. Moreover, the token-level replacement strategy, which only mods fewer words (typically five in the base setting), further ensures that the adversarial examples remain similar to the original context while fooling the model.

% Overall, the results demonstrate the superiority of our QA-Attack method in generating effective, high-quality adversarial examples. It outperforms baseline methods in disrupting model performance and maintaining contextual coherence, proving the efficacy of our approach in various question-answering scenarios.

\subsection{Ablation and Hyperparameters Studies} \label{Ablation Studies}
To comprehensively validate the efficacy of the proposed QA-Attack method, this section conducts a detailed ablation study, dissecting each component to assess its individual impact and overall contribution to the method’s performance.

\subsubsection{Effectiveness of Hybrid Fusion Ranking on Multiple Question Types}
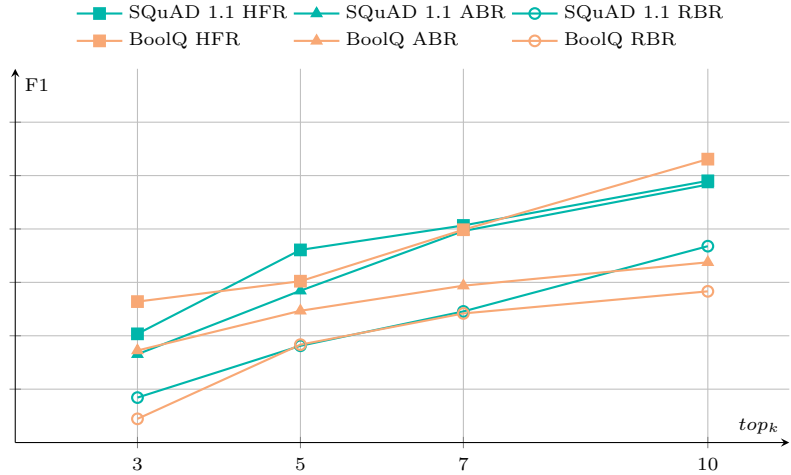
\begin{figure}[!t]
\centering
\begin{tikzpicture}
    \begin{axis}[
        width=0.9\textwidth,
        height=0.5\textwidth,
        xlabel={$top_k$},
        ylabel={F1},
        xmin=1.5, xmax=11,
        ymin=5, ymax=40,
        axis lines=center,
        grid=major,
        xtick={3, 5, 7, 10},
        ytick={0, 10, 15, 20, 25, 30, 35},
        legend style={at={(0.5,1.2)},legend columns=3,anchor=north,font=\footnotesize,draw=none},
        legend cell align={left},
        label style={font=\footnotesize},
    tick label style={font=\footnotesize}]
    ]
    % SQuAD lines with markers
    \addplot [thick, JungleGreen!80, mark=square*] table {
        0 nan
        3 15.18
        5 23.04
        7 25.33
        10 29.51
    
    };
    \addlegendentry{SQuAD 1.1 HFR}
    
    \addplot [thick, JungleGreen!80, mark=triangle*] table {
        0 nan
        3 13.27
        5 19.22
        7 24.81
        10 29.16
        13 nan
    };
    \addlegendentry{SQuAD 1.1 ABR}
    \addplot [thick, JungleGreen!80, mark=o, mark options={fill=JungleGreen!80}] table {
        0 nan
        3 9.21
        5 14.06
        7 17.28
        10 23.39
        13 nan
    };
    \addlegendentry{SQuAD 1.1 RBR}

    % BoolQ lines with markers
    \addplot [thick, Peach!80, draw=Peach!80, mark=square*] table {
        0 nan
        3 18.20
        5 20.11
        7 24.93
        10 31.54
        13 nan
    };
    \addlegendentry{BoolQ HFR}
    
    \addplot [thick, Peach!80, draw=Peach!80, mark=triangle*, mark options={fill=Peach!80}] table {
        0 nan
        3 13.62
        5 17.34
        7 19.69
        10 21.88
        13 nan
    };
    \addlegendentry{BoolQ ABR}

    \addplot [thick, Peach!80, draw=Peach!80, mark=o, mark options={fill=Peach!80}] table {
        0 nan
        3 7.22
        5 14.17
        7 17.09
        10 19.16
        13 nan
    };
    \addlegendentry{BoolQ RBR}
    \end{axis}
\end{tikzpicture}
\caption{F1 score analysis for HFR, ABR, and RBR variants of QA-Attack using different $top_k$ values, tested on datasets SQuAD 1.1 and BoolQ.}
\label{k_comparison}
\end{figure}

We test how HRF, ABR, and RBR methods perform across different $top_k$ values on the SQuAD and BoolQ datasets, with $d$ remaining, shown in Fig.~\ref{k_comparison}. HRF consistently outperforms ABR and RBR for all $top_k$ values on both datasets. This suggests combining attention-based and removal-based ranking in HRF is more effective at generating robust adversarial examples than using either method alone. The graph also shows that as $top_k$ increases, all methods improve, indicating that higher $top_k$ values help identify vulnerable tokens better and lead to more effective attacks.

Despite the better performance at higher $top_k$ values, the study uses $top_k = 5$ as a base setting. This choice balances effectiveness with minimal text modification, ensuring that adversarial examples remain close to the original context while still being effective. The consistent trend across both SQuAD and BoolQ datasets demonstrates that HRF's superior performance holds true for different question types, showing its versatility in attacking various question-answering models. This analysis highlights the practical effectiveness of the HRF method and its ability to generate impactful adversarial examples across different QA tasks.

\subsubsection{Effectiveness of Synonyms Selection}
To evaluate our Synonyms Selection approach, we conduct comparisons in two aspects. We first compare our BERT-based synonym generation against two alternative methods: WordNet \cite{WordNet}, an online database that contains sets of synonyms, and HowNet \cite{hownet}, which produces semantically similar words using its network structure. Using the base configuration, we evaluate the EM scores when attacking T5 and BERT\textsubscript{base} models across three datasets: SQuAD 1.1, NarrativeQA, and BoolQ. The results demonstrate that our QA-Attack with BERT\textsubscript{base} consistently achieved superior performance compared to other methods across all datasets and victim models.

\begin{table}[!t]
\centering
\caption{EM scores for attacks on T5 and BERT\textsubscript{base} models using three distinct synonym generation methods. Lower scores indicate more effective attacks.}
\begin{tabular}{p{2.5cm}p{2cm}p{1.5cm}p{2cm}p{1.5cm}}
\toprule
\multirow{2}{*}{Methods} & \multirow{2}{*}{Victim Models} & \multicolumn{3}{c}{Datasets}\\
\cmidrule(lr){3-5}
&& SQuAD 1.1 & NarrativeQA & BoolQ \\ 
\midrule
\multirow{2}{*}{HowNet} & T5 & 14.22 & 7.25 & 29.08 \\
& BERT\textsubscript{base} & 7.66 & 4.52 & 26.91\\
\multirow{2}{*}{WordNet} & T5 & 5.31 & 3.99 & 21.63\\
& BERT\textsubscript{base} & 7.23 & 5.67 & 19.35\\
\multirow{2}{*}{BERT\textsubscript{base} (ours)} & T5 & \textbf{4.67} & \textbf{5.61} & \textbf{8.64} \\
& BERT\textsubscript{base} & \textbf{6.42} & \textbf{3.86} & \textbf{18.39} \\ 
\bottomrule
\end{tabular}
\label{synonyms_selection}
\end{table}

On the other hand, we also examine the impact of parameter $d$ in Synonym Selection, which determines the number of synonyms obtained from the Masked Language Model (MLM). Table \ref{d performance} illustrates that as $d$ increases from 1 to 3, F1 scores consistently decrease across all datasets, indicating improved attack performance. This trend suggests that a more aggressive setting (higher $d$) is more effective in compromising model accuracy across various datasets.
\begin{table}[!t]
\centering
\caption{F1 scores demonstrating QA-Attack's performance across five datasets under different $d$ values (i.e., number of synonym candidates for substitutions).}
\label{d performance}
\begin{tabular}{p{1cm}p{1.5cm}p{2cm}p{1cm}p{1.5cm}p{1.5cm}}
\toprule
 & SQuAD 1.1 & SQuAD V2.0 & BoolQ & NarrativeQA & NewQA \\ 
\midrule
$d = 1$ & 8.52 & 14.72 & 19.22 & 7.63 & 10.66 \\ 
$d = 2$ & 4.67 & 9.13 & 15.16 & 5.61 & 3.61 \\ 
$d = 3$ & 2.17 & 7.26 & 11.43 & 3.71 & 3.27 \\
\bottomrule
\end{tabular}
\end{table}

\subsubsection{Texual Quality of Word Candidates}\label{Texual quality of the word candidates}
In our ablation study, detailed in Table \ref{candidate quality}, we investigate the quality of adversarial examples generated by various attack methods on the T5 model using the SQuAD 1.1 dataset. We evaluate our word replacement technique with encoder-decoder candidate generation (T3), as well as sentence-level modification methods (TASA, TMYC). The results indicate that our word-level synonym selection approach outperformed all other baselines. Notably, our word-level attack maintains a lower grammar error rate and higher linguistic fluency than alternative methods. Although RobustQA employs the same synonym selection strategy, it requires more word modifications to successfully attack the model and tends to produce more adventurous alterations.
\begin{table}[!t]
\centering
\caption{Performance metrics for different word candidate selection strategies against T5 model on SQuAD 1.1 dataset.}
\label{candidate quality}
\begin{tabular}{p{4.7cm}p{1cm}p{1cm}p{1cm}p{1cm}p{1cm}}
\toprule
Methods & EM$\downarrow$ & SIM$\uparrow$ & Mod$\downarrow$ & PPL$\downarrow$ & GErr$\downarrow$ \\
\midrule
TASA~\cite{TASA} & 9.21 & 6.38 & 8.15 & 143 & 0.13 \\
TMYC (Tick Me If You Can)~\cite{wallace2019trick} & 7.28 & 8.22 & 9.21 & 151 & 0.14 \\
RobustQA~\cite{robustmultilingual} & 5.89 & 6.03 & 8.35 & 147 & 0.15 \\
T3~\cite{t3} & 5.41 & 7.23 & 7.93 & 133 & 0.13 \\
TextFooler~\cite{cla_att_random} & 10.60 & 6.29 & 8.17 & 136 & 0.14 \\
QA-Attack (ours) & \textbf{5.68} & \textbf{5.91} & \textbf{7.24} & \textbf{125} & \textbf{0.12} \\ 
\bottomrule
\end{tabular}
\end{table}

\subsection{Platform and Efficiency Analysis} \label{efficiency}
In this section, we evaluate QA-Attack's computational efficiency under base settings. We measure efficiency using time consumption per sample, expressed in seconds, where a lower value indicates superior performance. As shown in Table \ref{Time consumption}, the outcomes reveal that QA-Attack exhibits remarkable time efficiency, consistently outperforming baseline methods across both long-text (NarrativeQA) and short-text (SQuAD 1.1) datasets. This superior performance can be attributed to QA-Attack's innovative Hybrid Ranking Fusion (HRF) strategy, which effectively identifies vulnerable words within the text, significantly enhancing the speed of the attack process.

\begin{table}[!t]
\centering
\caption{Time consumption (seconds per sample) for various methods and datasets. A lower value indicates better performance.}
\label{Time consumption}
\begin{tabular}{p{4.3cm}p{1.4cm}p{1.5cm}p{1.7cm}p{0.8cm}p{0.6cm}}
\toprule
 & NarrativeQA & SQuAD 1.1 & SQuAD V2.0 & NewsQA & BoolQ \\ 
\midrule
TASA~\cite{TASA} & 28.77 & 15.82 & 18.25 & 10.72 & -- \\
TMYC (Tick Me If You Can)~\cite{wallace2019trick} & 25.61 & 12.75 & 16.33 & 9.21 & 7.42 \\
RobustQA~\cite{robustmultilingual} & 25.82 & 24.46 & 22.15 & 12.81 & 15.82 \\
T3~\cite{t3} & 26.52 & 21.37 & 28.38 & 14.74 & 7.93 \\
QA-Attack (ours) & \textbf{23.51} & \textbf{10.61} & \textbf{12.38} & \textbf{8.32} & \textbf{7.22} \\ 
\bottomrule
\end{tabular}
\end{table}

\subsection{Adversarial Retraining}\label{Adversarial Retraining}
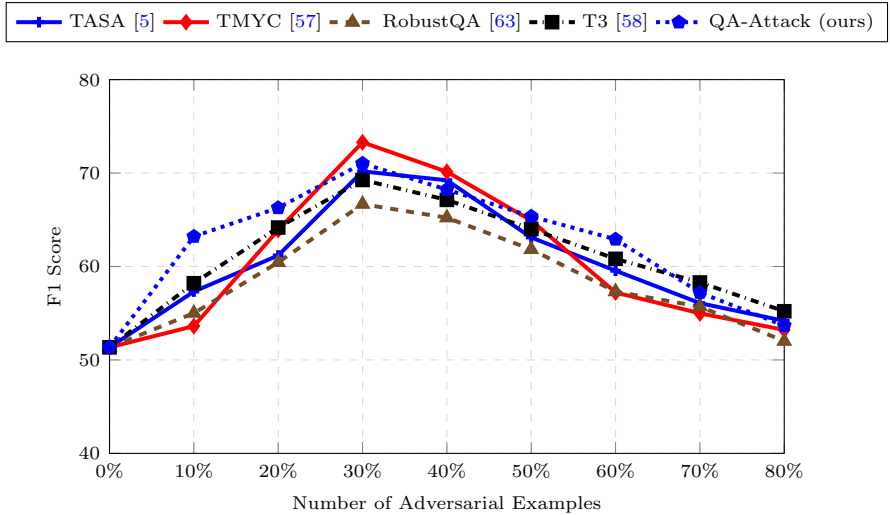
\begin{figure}[!t]
\centering
\begin{tikzpicture}
\begin{axis}[
    width=0.8\textwidth, height=0.5\textwidth,
    xlabel={Number of Adversarial Examples}, % Replace with actual x-axis label if needed
    ylabel={F1 Score},
    xmin=0, xmax=80,
    ymin=40, ymax=80,
    xtick={0, 10, 20, 30, 40, 50, 60, 70, 80},
    ytick={40, 50, 60, 70, 80},
    xticklabels={0\%, 10\%, 20\%, 30\%, 40\%, 50\%, 60\%, 70\%, 80\%},
    yticklabels={40, 50, 60, 70, 80},
    grid=both,
    grid style={dashed, gray!30},
    legend style={at={(0.5,1.1)}, anchor=south,legend columns=-1,font=\footnotesize},
    legend cell align={left},
    mark options={solid},
    label style={font=\footnotesize},
    tick label style={font=\footnotesize} 
]

% TASA
\addplot+[mark=+, line width=1.5pt] coordinates {(0, 51.35) (10, 57.32) (20, 61.18) (30, 70.18) (40, 69.23) (50, 63.11) (60, 59.55) (70, 56.08) (80, 54.18)};
\addlegendentry{TASA~\cite{TASA}}

% TMYC
\addplot+[mark=diamond*, line width=1.5pt] coordinates {(0, 51.35) (10, 53.62) (20, 63.91) (30, 73.29) (40, 70.14) (50, 64.92) (60, 57.22) (70, 54.99) (80, 53.21)};
\addlegendentry{TMYC~\cite{wallace2019trick}}

% RobustQA
\addplot+[mark=triangle*, dashed, line width=1.5pt] coordinates {(0, 51.35) (10, 55.01) (20, 60.44) (30, 66.67) (40, 65.22) (50, 61.83) (60, 57.31) (70, 55.71) (80, 52.01)};
\addlegendentry{RobustQA~\cite{robustmultilingual}}

% T3
\addplot+[mark=square*, dash dot, line width=1.5pt] coordinates {(0, 51.35) (10, 58.21) (20, 64.17) (30, 69.26) (40, 67.13) (50, 64.01) (60, 60.82) (70, 58.29) (80, 55.21)};
\addlegendentry{T3~\cite{t3}}

% QA-Attack
\addplot+[mark=pentagon*, dotted, line width=1.5pt] coordinates {(0, 51.35) (10, 63.21) (20, 66.28) (30, 71.03) (40, 68.23) (50, 65.36) (60, 62.91) (70, 57.21) (80, 53.67)};
\addlegendentry{QA-Attack (ours)}
\end{axis}
\end{tikzpicture}
\caption{The performance of the T5 model re-trained on the SQuAD 1.1 dataset with mixed adversarial examples generated by TASA~\cite{TASA}, TMYC~\cite{wallace2019trick}, RobustQA~\cite{robustmultilingual}, T3~\cite{t3}, and our QA-Attack.}
\label{retraining}
\end{figure}

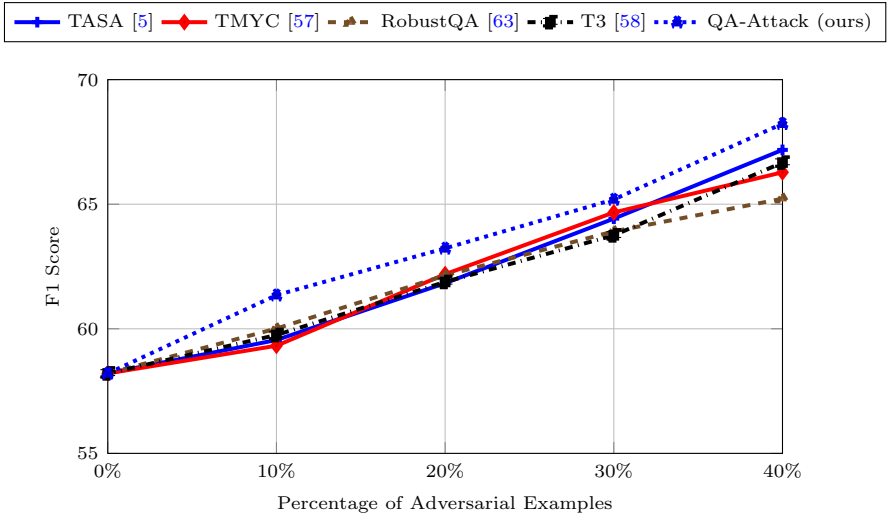
\begin{figure}[!t]
\centering
\begin{tikzpicture}
\begin{axis}[
    width=0.8\textwidth, height=0.5\textwidth,
    xlabel={Percentage of Adversarial Examples},
    ylabel={F1 Score},
    xmin=0, xmax=40,
    ymin=55, ymax=70,
    xtick={0, 10, 20, 30, 40},
    ytick={55, 60, 65, 70},
    xticklabels={0\%, 10\%, 20\%, 30\%, 40\%},
    yticklabels={55, 60, 65, 70},
    grid=both,
    legend style={at={(0.5,1.1)}, anchor=south, legend columns=-1, font=\footnotesize},
    legend cell align={left},
    label style={font=\footnotesize},
    tick label style={font=\footnotesize} 
]

% TASA
\addplot+[mark=+, line width=1.5pt] coordinates {(0, 58.21) (10, 59.55) (20, 61.83) (30, 64.42) (40, 67.18)};
\addlegendentry{TASA~\cite{TASA}}

% TMYC
\addplot+[mark=diamond*, line width=1.5pt] coordinates {(0, 58.21) (10, 59.32) (20, 62.19) (30, 64.67) (40, 66.28)};
\addlegendentry{TMYC~\cite{wallace2019trick}}

% RobustQA
\addplot+[mark=triangle*, dashed, line width=1.5pt] coordinates {(0, 58.21) (10, 60.01) (20, 62.14) (30, 63.91) (40, 65.22)};
\addlegendentry{RobustQA~\cite{robustmultilingual}}

% T3
\addplot+[mark=square*, dash dot, line width=1.5pt] coordinates {(0, 58.21) (10, 59.75) (20, 61.88) (30, 63.75) (40, 66.67)};
\addlegendentry{T3~\cite{t3}}

% QA-Attack
\addplot+[mark=pentagon*, dotted, line width=1.5pt] coordinates {(0, 58.21) (10, 61.35) (20, 63.23) (30, 65.19) (40, 68.23)};
\addlegendentry{QA-Attack (ours)}

\end{axis}
\end{tikzpicture}
\caption{F1 scores of attacking T5 models retrained with increasing proportions of adversarial examples generated by baseline methods (TASA~\cite{TASA}, TMYC~\cite{wallace2019trick}, RobustQA~\cite{robustmultilingual}, T3~\cite{t3}) and our QA-Attack.}
\label{fig:attack_retraining}
\end{figure}

In this section, we investigate QA-Attack's potential for enhancing downstream models' accuracy. We employ QA-Attack to generate adversarial examples from SQuAD 1.1 training sets and incorporate them as supplementary training data. We reconstruct the training set with varying proportions of adversarial examples added to the raw training set. The retraining process with this augmented data aims to examine how test accuracy changes in response to the inclusion of adversarial examples. As illustrated in Fig.~\ref{retraining}, re-training with adversarial examples slightly improves model performance when less than 30\% of the training data consists of adversaries. However, performance decreases when the proportion of adversaries exceeds 30\%. This finding indicates that the optimal ratio of adversarial examples in training data needs to be determined empirically, which aligns with conclusions from previous attacking methods. To evaluate how re-training helps defend against adversarial attacks, we analyze the robustness of T5 models trained with varying proportions of adversarial examples (0\%, 10\%, 20\%, 30\%, 40\%) from different attack methods, as shown in Fig.~\ref{retraining}. A lower F1 score indicates higher model susceptibility to adversarial attacks. The attack performance of the re-trained model is shown in Fig.~\ref{fig:attack_retraining}. It demonstrates that incorporating adversarial examples during training consistently improves model robustness, as evidenced by increasing F1 scores across all attack methods. Notably, QA-Attack emerges as the most effective approach, consistently outperforming other methods, with its advantage becoming particularly pronounced at higher percentages of adversarial training data.

\subsection{Attacking Models with Defense Mechanism}\label{defense}
Defending NLP models against adversarial attacks is crucial for maintaining the reliability of language processing systems in real-world applications \cite{defense}. To further analyse how attacks are performed under defense systems, we deploy two distinct defense mechanisms to investigate our attack performance under defense systems. The first is Frequency-Guided Word Substitutions (FGWS) approach \cite{FGWS}, which excels at detecting adversarial examples. The second is Random Masking Training (RanMASK) \cite{RanMASK}, a technique that enhances model robustness through specialized training procedures. We perform the adversarial attack on T5 on datasets SQuAD 1.1, NarrativeQA and BoolQ, and the results are presented in Table \ref{tab:defensive_analysis}. The results show that QA-Attack demonstrates superior adversarial robustness across multiple benchmark datasets, consistently outperforming existing methods against state-of-the-art defenses.

\begin{table}[!t]
\centering
\caption{Effectiveness of defense mechanisms (FGWS~\cite{FGWS} and RanMASK~\cite{RanMASK}) against QA-Attack: EM scores of T5 model output answers across SQuAD 1.1, NarrativeQA, and BoolQ datasets. Lower scores indicate higher attack success against defenses.}
\begin{tabular}{llccccc}
\toprule
Datasets & Defense & TASA & RobustQA & TMYC & T3 & QA-Attack \\ 
\midrule
\multirow{2}*{SQuAD 1.1} & FGWS~\cite{FGWS} & 34.71 & 39.42 & 28.51 & 24.11 & \textbf{21.03} \\ 
&RanMASK~\cite{RanMASK} & 32.17 & 39.78 & 44.81 & 41.09 & \textbf{30.26} \\ 
\multirow{2}*{NarrativeQA} & FGWS~\cite{FGWS} & 49.28 & 44.62 & \textbf{37.21} & 45.17 & 38.33 \\
& RanMASK~\cite{RanMASK} & 38.41 & 37.14 & 41.62 & 43.81 & \textbf{34.47} \\ 
\multirow{2}*{BoolQ} & FGWS~\cite{FGWS} & 45.71 & 47.37 & 38.97 & 45.33 & \textbf{38.34}\\
& RanMASK~\cite{RanMASK} & 41.63 & 42.88 & 47.25 & 42.17 & \textbf{40.51}\\
\bottomrule
\end{tabular}
\label{tab:defensive_analysis}
\end{table}

\subsection{Transferability of Attacks} \label{Transferability of Attacks}
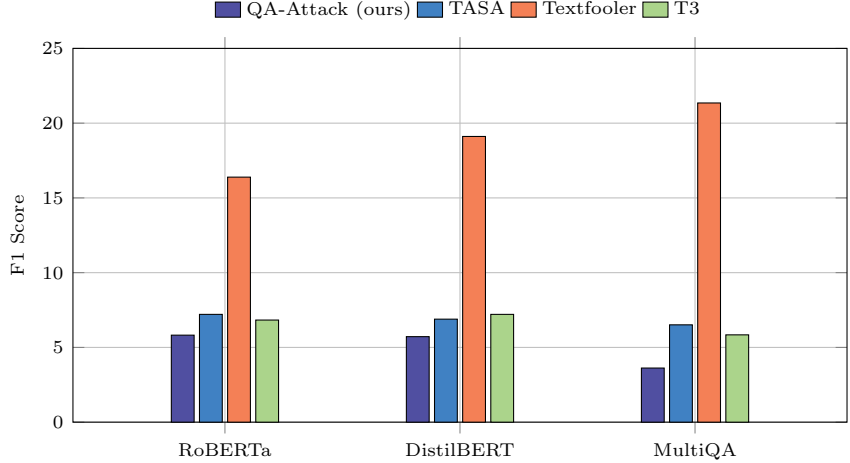
\begin{figure}[!t]
    \centering
    \begin{tikzpicture}
    \begin{axis}[
        ybar,
        ylabel={F1 Score},
        enlarge x limits={abs=2cm},
        xtick=data,
        grid=major,
        symbolic x coords={RoBERTa, DistilBERT, MultiQA},
        ymin= 0, ymax= 25,
        bar width=0.3cm,
        width=0.9\textwidth,
        height=0.5\textwidth,
        legend style={at={(0.5,1.05)},
        anchor=south,legend columns=-1, draw=none}, 
        legend image code/.code={%
                        \draw[#1, draw=black] (0cm,-0.1cm) rectangle (0.33cm,0.1cm);
                    },
        legend entries={QA-Attack (ours), TASA, Textfooler, T3},
        yticklabel style={
            /pgf/number format/fixed,       % Use fixed-point notation
            },
        legend style={font=\footnotesize},
        label style={font=\footnotesize},
        tick label style={font=\footnotesize} 
    ]
    \addplot[fill=Blue!80, draw=black] coordinates {(RoBERTa,5.82) (DistilBERT,5.72) (MultiQA,3.62)};
    \addplot[fill=NavyBlue!80, draw=black] coordinates {(RoBERTa,7.21) (DistilBERT,6.89) (MultiQA,6.51)};
    \addplot[fill=RedOrange!80, draw=black] coordinates {(RoBERTa,16.39) (DistilBERT,19.11) (MultiQA,21.35)};
    \addplot[fill=YellowGreen!80, draw=black] coordinates {(RoBERTa,6.83) (DistilBERT,7.21) (MultiQA,5.84)};
    \end{axis}
    \end{tikzpicture}
    \caption{F1 scores for transfer attacks on three other QA models using adversarial samples generated for UnifiedQA. A lower value indicates better performance.}
    \label{transfering_attack_f1}
\end{figure}

To evaluate our model's transferability, we test the adversarial samples generated for T5 on three distinct question-answering models: RoBERTa \cite{RoBERTa}, DistilBERT \cite{DistilBERT}, and MultiQA \cite{MultiQA}. We also compare the transferability of three baseline methods: TASA, TextFooler, and T3, under identical experimental conditions. As shown in Fig.~\ref{transfering_attack_f1}, QA-Attack effectively degrades other QA models' performance on both the NarrativeQA and BoolQ datasets. This suggests that the transferring attack performance of our QA-Attack consistently outperforms the baselines.
% The results of transferring adversarial examples from the UnifiedQA model to the BERT-base model reveal that QA-Attack achieves the best transferability among all attack strategies. The low F1 and EM scores indicate a significant degradation in the BERT-base's performance, while the high ROUGE1 and BLEU1 scores confirm that the adversarial examples remain contextually coherent. Additionally, the lower similarity scores further validate the effectiveness of QA-Attack in introducing substantial semantic differences in BERT-base's predictions.
\subsection{Parts of Speech Preference} \label{Parts of speech preference}
To further understand the candidate words' distribution of our word-level attack, we examine its attacking preference in terms of Parts of Speech (POS), highlighting vulnerable areas within the input context. We use the Stanford POS tagger \cite{tagger} to label each attacked word, categorizing them as \textit{noun}, \textit{verb}, \textit{adjective (Adj.)}, \textit{adverb (Adv.)}, and \textit{others (e.g., pronoun, preposition, conjunction)}. Table \ref{speech preference} illustrates the POS preference of our QA-Attack compared to baseline methods in the base setting. For ``informative queries'' on SQuAD dataset, most attacking methods predominantly target \textit{nouns}, while TASA shows a slight preference for \textit{adverbs}. In the case of ``boolean queries'' on BoolQ dataset, all methods frequently focus on \textit{adjectives} and \textit{adverbs}. Notably, our QA-Attack demonstrates a higher preference for the ``others'' category. Given that these parts of speech (pronouns, prepositions, and conjunctions) carry limited semantic content, we suggest that altering them may not significantly affect the linguistic or semantic aspects of prediction. However, such modifications could disrupt sequential dependencies, potentially compromising the contextual understanding of QA models and misleading their answers.
\begin{table}[!t]
\centering
\caption{POS preference with respect to choices of victim words among attacking methods.}
\begin{tabular}{llccccc}
\toprule
Datasets & Methods & Noun (\%) & Verb (\%) & Adj. (\%) & Adv. (\%) & Others (\%) \\ 
\midrule
\multirow{6}*{SQuAD 1.1} & TASA & \textit{35} & 12 & 13 & \textbf{36} & 4 \\
&TMYC & \textbf{47} & \textit{21} & 11 & 5 & 17 \\
&RobustQA & \textbf{34} & 13 & \textit{22} & 16 & 15 \\
&Textfooler & \textbf{44} & 13 & \textit{23} & 8 & 12 \\
&T3 & \textbf{60} & \textit{17} & 6 & 7 & 10 \\
&QA-Attack & \textit{34} & 9 & 18 & 3 & \textbf{36} \\
\midrule
\multirow{6}*{BoolQ} & TASA & -- & -- & - & -- & - \\
&TMYC & 14 & \textit{19} & 12 & \textbf{35} & 20 \\
&RobustQA & \textit{19} & 14 & \textbf{27} & 23 & 17 \\
&Textfooler & \textbf{41} & 15 & \textit{27} & 7 & 10 \\
&T3 & \textbf{42} & \textit{13} & 20 & 16 & 9 \\
&QA-Attack & 10 & 19 & \textit{25} & 18 & \textbf{28} \\
\bottomrule
\end{tabular}
\label{speech preference}
\end{table}

\subsection{Robustness versus the Scale of Pre‑trained Models}\label{Robustness versus the scale of pre‑trained models}
From the attacking results in Table~\ref{Bert_results} discussed in Sec~\ref{experiment analysis}, we recognize the limitation of our QA-Attack on BERT\textsubscript{base}, with $L=12$ and $H=768$, which does not sufficiently support robust experimental outcomes. To address this issue and gain more comprehensive insights, we conducted experiments with four different sizes of BERT \cite{bert} models\footnote{Different sizes of BERT models can be obtained from \url{https://github.com/google-research/bert/}}: BERT\textsubscript{tiny}, BERT\textsubscript{mini}, BERT\textsubscript{medium}, and BERT\textsubscript{large}. Our findings, detailed in Table \ref{model size}, demonstrate a positive correlation between model size and experimental robustness. The effectiveness of adversarial attacks decreases as the complexity and capacity of the BERT model increase, suggesting that deeper architectures provide better protection against adversarial perturbations.

\begin{table}[!t]
\centering
\caption{A comparative analysis to attacking various sizes of BERT model on SQuAD 1.1 dataset. A lower value indicates better attack performance.}
\begin{tabular}{ccccc}
\toprule
Versions & BERT tiny & BERT mini & BERT medium & BERT large\\ 
\midrule
Size & L = 2, H = 128 & L = 4, H = 256 & L = 8, H = 512 & L = 24, H = 1024 \\ 
\midrule
EM $\downarrow$ & 11.82 & 13.26 & 13.31 & 14.25 \\
F1 $\downarrow$ & 5.67 & 6.35 & 6.42 & 7.24 \\
SIM $\downarrow$ & 6.23 & 7.12 & 7.43 & 8.38 \\
\bottomrule
\end{tabular}
\label{model size}
\end{table}

\section{Conclusion and Future Work}\label{Conclusion and Future Work}
% In conclusion, QA-Attack represents a significant advancement in adversarial attacks on question-answering models. Our Hybrid Ranking Fusion approach effectively identifies vulnerable tokens, outperforming existing methods in attack success rate and efficiency while maintaining linguistic coherence. The method's versatility across different question types demonstrates its robustness and real-world potential. Future work will explore transferability across model architectures and investigate defence mechanisms, contributing to developing more secure AI systems in natural language processing.

The robustness of QA models has been increasingly challenged by adversarial attacks. These attacks expose the vulnerabilities of models used in various tasks, including information retrieval, conversational agents, and machine comprehension. To address this, we introduced QA-Attack, which leverages Hybrid Ranking Fusion (HRF) to conduct effective attacks by identifying and modifying the most critical tokens in the input text. Through a combination of attention-based and removal-based ranking strategies, QA-Attack successfully disrupts model predictions while maintaining high levels of semantic and linguistic coherence. Extensive experiments have demonstrated that our method outperforms existing attack techniques regarding attack success, fluency, and consumption across various datasets, confirming its efficacy in undermining the robustness of state-of-the-art QA models.

While adversarial attacks such as QA-Attack highlight the weaknesses in QA systems, they also provide an opportunity to test and improve model robustness. In future work, we plan to focus on developing defence strategies that mitigate these vulnerabilities. Furthermore, we intend to extend QA-Attack to handle more complex and diverse QA scenarios, including multiple-choice questions and multi-hop reasoning \cite{multihop}, to ensure that our method remains a powerful tool for evaluating and improving the robustness of QA systems in an evolving landscape of adversarial threats.

\bibliographystyle{splncs04}

\bibliography{bibliography.bib}

\begin{thebibliography}{10}
\providecommand{\url}[1]{\texttt{#1}}
\providecommand{\urlprefix}{URL }
\providecommand{\doi}[1]{https://doi.org/#1}

\bibitem{gpt4}
Antaki, F., Milad, D., Chia, M.A., Gigu{\`e}re, C.{\'E}., Touma, S., El-Khoury, J., Keane, P.A., Duval, R.: Capabilities of gpt-4 in ophthalmology: an analysis of model entropy and progress towards human-level medical question answering. British Journal of Ophthalmology  \textbf{108}(10),  1371--1378 (2024)

\bibitem{att-translate}
Bahdanau, D., Cho, K., Bengio, Y.: Neural machine translation by jointly learning to align and translate. CoRR  \textbf{abs/1409.0473} (2014)

\bibitem{gpt3}
Bongini, P., Becattini, F., Del~Bimbo, A.: Is gpt-3 all you need for visual question answering in cultural heritage? In: European Conference on Computer Vision. pp. 268--281. Springer (2022)

\bibitem{gpt}
Brown, T., Mann, B., Ryder, N., Subbiah, M., Kaplan, J.D., Dhariwal, P., Neelakantan, A., Shyam, P., Sastry, G., Askell, A., Agarwal, S., Herbert-Voss, A., Krueger, G., Henighan, T., Child, R., Ramesh, A., Ziegler, D., Wu, J., Winter, C., Hesse, C., Chen, M., Sigler, E., Litwin, M., Gray, S., Chess, B., Clark, J., Berner, C., McCandlish, S., Radford, A., Sutskever, I., Amodei, D.: Language models are few-shot learners. In: Larochelle, H., Ranzato, M., Hadsell, R., Balcan, M., Lin, H. (eds.) Advances in Neural Information Processing Systems. vol.~33, pp. 1877--1901. Curran Associates, Inc. (2020)

\bibitem{TASA}
Cao, Y., Li, D., Fang, M., Zhou, T., Gao, J., Zhan, Y., Tao, D.: {TASA}: Deceiving question answering models by twin answer sentences attack. In: Goldberg, Y., Kozareva, Z., Zhang, Y. (eds.) Proceedings of the 2022 Conference on Empirical Methods in Natural Language Processing. pp. 11975--11992. Association for Computational Linguistics, Abu Dhabi, United Arab Emirates (Dec 2022)

\bibitem{BoolQ}
Clark, C., Lee, K., Chang, M.W., Kwiatkowski, T., Collins, M., Toutanova, K.: {B}ool{Q}: Exploring the surprising difficulty of natural yes/no questions. In: Burstein, J., Doran, C., Solorio, T. (eds.) Proceedings of the 2019 Conference of the North {A}merican Chapter of the Association for Computational Linguistics: Human Language Technologies, Volume 1 (Long and Short Papers). pp. 2924--2936. Association for Computational Linguistics, Minneapolis, Minnesota (Jun 2019)

\bibitem{bert}
Devlin, J., Chang, M.W., Lee, K., Toutanova, K.: {BERT}: Pre-training of deep bidirectional transformers for language understanding. In: Proceedings of the 2019 Conference of the North American Chapter of the Association for Computational Linguistics: Human Language Technologies. pp. 4171--4186. Association for Computational Linguistics, Minneapolis, Minnesota (June 2019)

\bibitem{dong2022adversarial}
Dong, H., Dong, J., Yuan, S., Guan, Z.: Adversarial attack and defense on natural language processing in deep learning: A survey and perspective. In: International conference on machine learning for cyber security. pp. 409--424. Springer (2022)

\bibitem{hownet}
Dong, Z., Dong, Q.: Hownet-a hybrid language and knowledge resource. In: International conference on natural language processing and knowledge engineering, 2003. Proceedings. 2003. pp. 820--824. IEEE (2003)

\bibitem{hotflip}
Ebrahimi, J., Rao, A., Lowd, D., Dou, D.: {H}ot{F}lip: White-box adversarial examples for text classification. In: Gurevych, I., Miyao, Y. (eds.) Proceedings of the 56th Annual Meeting of the Association for Computational Linguistics (Volume 2: Short Papers). pp. 31--36. Association for Computational Linguistics, Melbourne, Australia (Jul 2018)

\bibitem{attention_review}
Galassi, A., Lippi, M., Torroni, P.: Attention in natural language processing. IEEE Transactions on Neural Networks and Learning Systems  \textbf{32}(10),  4291–4308 (Oct 2021)

\bibitem{bae}
Garg, S., Ramakrishnan, G.: Bae: Bert-based adversarial examples for text classification. In: Proceedings of the 2020 Conference on Empirical Methods in Natural Language Processing (EMNLP). Association for Computational Linguistics (2020)

\bibitem{defense}
Goyal, S., Doddapaneni, S., Khapra, M.M., Ravindran, B.: A survey of adversarial defenses and robustness in nlp. ACM Computing Surveys  \textbf{55}(14s),  1--39 (2023)

\bibitem{LongT5}
Guo, M., Ainslie, J., Uthus, D., Ontanon, S., Ni, J., Sung, Y.H., Yang, Y.: {L}ong{T}5: {E}fficient text-to-text transformer for long sequences. In: Carpuat, M., de~Marneffe, M.C., Meza~Ruiz, I.V. (eds.) Findings of the Association for Computational Linguistics: NAACL 2022. pp. 724--736. Association for Computational Linguistics, Seattle, United States (Jul 2022)

\bibitem{hathaliya2022adversarial}
Hathaliya, J.J., Tanwar, S., Sharma, P.: Adversarial learning techniques for security and privacy preservation: A comprehensive review. Security and Privacy  \textbf{5}(3), ~e209 (2022)

\bibitem{iyyer-etal-2018-adversarial}
Iyyer, M., Wieting, J., Gimpel, K., Zettlemoyer, L.: Adversarial example generation with syntactically controlled paraphrase networks. In: Walker, M., Ji, H., Stent, A. (eds.) Proceedings of the 2018 Conference of the North {A}merican Chapter of the Association for Computational Linguistics: Human Language Technologies, Volume 1 (Long Papers). pp. 1875--1885. Association for Computational Linguistics, New Orleans, Louisiana (Jun 2018)

\bibitem{jia-liang-2017-adversarial}
Jia, R., Liang, P.: Adversarial examples for evaluating reading comprehension systems. In: Palmer, M., Hwa, R., Riedel, S. (eds.) Proceedings of the 2017 Conference on Empirical Methods in Natural Language Processing. pp. 2021--2031. Association for Computational Linguistics, Copenhagen, Denmark (Sep 2017)

\bibitem{cla_att_random}
Jin, D., Jin, Z., Zhou, J.T., Szolovits, P.: Is bert really robust? a strong baseline for natural language attack on text classification and entailment. Proceedings of the AAAI Conference on Artificial Intelligence  \textbf{34}(05),  8018--8025 (Apr 2020)

\bibitem{medicine}
Jin, D., Pan, E., Oufattole, N., Weng, W.H., Fang, H., Szolovits, P.: What disease does this patient have? a large-scale open domain question answering dataset from medical exams. Applied Sciences  \textbf{11}(14), ~6421 (2021)

\bibitem{kann-etal-2018-sentence}
Kann, K., Rothe, S., Filippova, K.: Sentence-level fluency evaluation: References help, but can be spared! In: Korhonen, A., Titov, I. (eds.) Proceedings of the 22nd Conference on Computational Natural Language Learning. pp. 313--323. Association for Computational Linguistics, Brussels, Belgium (Oct 2018)

\bibitem{UnifiedQA}
Khashabi, D., Min, S., Khot, T., Sabharwal, A., Tafjord, O., Clark, P., Hajishirzi, H.: {UNIFIEDQA}: Crossing format boundaries with a single {QA} system. In: Cohn, T., He, Y., Liu, Y. (eds.) Findings of the Association for Computational Linguistics: EMNLP 2020. pp. 1896--1907. Association for Computational Linguistics, Online (Nov 2020)

\bibitem{gpt2}
Klein, T., Nabi, M.: Learning to answer by learning to ask: Getting the best of gpt-2 and bert worlds. arXiv preprint arXiv:1911.02365  (2019)

\bibitem{qa_usage}
Klopfenstein, L.C., Delpriori, S., Malatini, S., Bogliolo, A.: The rise of bots: A survey of conversational interfaces, patterns, and paradigms. In: Proceedings of the 2017 Conference on Designing Interactive Systems. p. 555–565. DIS '17, Association for Computing Machinery, New York, NY, USA (2017)

\bibitem{ko-etal-2020-look}
Ko, M., Lee, J., Kim, H., Kim, G., Kang, J.: Look at the first sentence: Position bias in question answering. In: Webber, B., Cohn, T., He, Y., Liu, Y. (eds.) Proceedings of the 2020 Conference on Empirical Methods in Natural Language Processing (EMNLP). pp. 1109--1121. Association for Computational Linguistics, Online (Nov 2020)

\bibitem{NarrativeQA}
Ko{\v{c}}isk{\'y}, T., Schwarz, J., Blunsom, P., Dyer, C., Hermann, K.M., Melis, G., Grefenstette, E.: The {N}arrative{QA} reading comprehension challenge. Transactions of the Association for Computational Linguistics  \textbf{6},  317--328 (2018)

\bibitem{ALBERT}
Lan, Z., Chen, M., Goodman, S., Gimpel, K., Sharma, P., Soricut, R.: Albert: A lite bert for self-supervised learning of language representations. In: International Conference on Learning Representations (2020)

\bibitem{BART}
Lewis, M., Liu, Y., Goyal, N., Ghazvininejad, M., Mohamed, A., Levy, O., Stoyanov, V., Zettlemoyer, L.: {BART}: Denoising sequence-to-sequence pre-training for natural language generation, translation, and comprehension. In: Jurafsky, D., Chai, J., Schluter, N., Tetreault, J. (eds.) Proceedings of the 58th Annual Meeting of the Association for Computational Linguistics. pp. 7871--7880. Association for Computational Linguistics, Online (Jul 2020)

\bibitem{attack_nlp}
Li, D., Zhang, Y., Peng, H., Chen, L., Brockett, C., Sun, M.T., Dolan, B.: Contextualized perturbation for textual adversarial attack. In: Toutanova, K., Rumshisky, A., Zettlemoyer, L., Hakkani-Tur, D., Beltagy, I., Bethard, S., Cotterell, R., Chakraborty, T., Zhou, Y. (eds.) Proceedings of the 2021 Conference of the North American Chapter of the Association for Computational Linguistics: Human Language Technologies. pp. 5053--5069. Association for Computational Linguistics, Online (Jun 2021)

\bibitem{TextBugger}
Li, J., Ji, S., Du, T., Li, B., Wang, T.: Textbugger: Generating adversarial text against real-world applications. In: Proceedings 2019 Network and Distributed System Security Symposium. NDSS 2019, Internet Society (2019)

\bibitem{summarization_attack}
Li, J., Liu, W.: Summarization attack via paraphrasing (student abstract). In: Proceedings of the AAAI Conference on Artificial Intelligence. vol.~37, pp. 16250--16251 (2023)

\bibitem{bert_attack}
Li, L., Ma, R., Guo, Q., Xue, X., Qiu, X.: {BERT}-{ATTACK}: Adversarial attack against {BERT} using {BERT}. In: Webber, B., Cohn, T., He, Y., Liu, Y. (eds.) Proceedings of the 2020 Conference on Empirical Methods in Natural Language Processing (EMNLP). pp. 6193--6202. Association for Computational Linguistics, Online (Nov 2020)

\bibitem{dtccf}
Liang, B., Li, H., Su, M., Bian, P., Li, X., Shi, W.: Deep text classification can be fooled. In: Proceedings of the Twenty-Seventh International Joint Conference on Artificial Intelligence. p. 4208–4215. International Joint Conferences on Artificial Intelligence Organization (Jul 2018)

\bibitem{rouge}
Lin, C.Y.: Rouge: A package for automatic evaluation of summaries. In: Text summarization branches out. pp. 74--81 (2004)

\bibitem{attention_translation}
Luong, T., Pham, H., Manning, C.D.: Effective approaches to attention-based neural machine translation. In: M{\`a}rquez, L., Callison-Burch, C., Su, J. (eds.) Proceedings of the 2015 Conference on Empirical Methods in Natural Language Processing. pp. 1412--1421. Association for Computational Linguistics, Lisbon, Portugal (Sep 2015)

\bibitem{legal}
Martinez-Gil, J.: A survey on legal question--answering systems. Computer Science Review  \textbf{48},  100552 (2023)

\bibitem{word2vec}
Mikolov, T., Chen, K., Corrado, G., Dean, J.: Efficient estimation of word representations in vector space. In: Proceedings of the 1st International Conference on Learning Representations (ICLR) (2013)

\bibitem{WordNet}
Miller, G.A.: {W}ord{N}et: A lexical database for {E}nglish. In: Speech and Natural Language: Proceedings of a Workshop Held at Harriman, New York, {F}ebruary 23-26, 1992 (1992)

\bibitem{FGWS}
Mozes, M., Stenetorp, P., Kleinberg, B., Griffin, L.: Frequency-guided word substitutions for detecting textual adversarial examples. In: Proceedings of the 16th Conference of the European Chapter of the Association for Computational Linguistics: Main Volume. pp. 171--186. Association for Computational Linguistics, Online (Apr 2021)

\bibitem{grammar}
Naber, D.: A Rule-Based Style and Grammar Checker. GRIN Verlag (2003)

\bibitem{chatbot}
Nuruzzaman, M., Hussain, O.K.: A survey on chatbot implementation in customer service industry through deep neural networks. In: 2018 IEEE 15th International Conference on e-Business Engineering (ICEBE). pp. 54--61 (2018)

\bibitem{bleu}
Papineni, K., Roukos, S., Ward, T., Zhu, W.J.: Bleu: a method for automatic evaluation of machine translation. In: Proceedings of the 40th annual meeting of the Association for Computational Linguistics. pp. 311--318 (2002)

\bibitem{radford2019language}
Radford, A., Wu, J., Child, R., Luan, D., Amodei, D., Sutskever, I., et~al.: Language models are unsupervised multitask learners. OpenAI blog  \textbf{1}(8), ~9 (2019)

\bibitem{T5}
Raffel, C., Shazeer, N., Roberts, A., Lee, K., Narang, S., Matena, M., Zhou, Y., Li, W., Liu, P.J.: Exploring the limits of transfer learning with a unified text-to-text transformer. Journal of machine learning research  \textbf{21}(140),  1--67 (2020)

\bibitem{SQuADv2}
Rajpurkar, P., Jia, R., Liang, P.: Know what you don{'}t know: Unanswerable questions for {SQ}u{AD}. In: Gurevych, I., Miyao, Y. (eds.) Proceedings of the 56th Annual Meeting of the Association for Computational Linguistics (Volume 2: Short Papers). pp. 784--789. Association for Computational Linguistics, Melbourne, Australia (Jul 2018)

\bibitem{SQuAD1.1}
Rajpurkar, P., Zhang, J., Lopyrev, K., Liang, P.: {SQ}u{AD}: 100,000+ questions for machine comprehension of text. In: Su, J., Duh, K., Carreras, X. (eds.) Proceedings of the 2016 Conference on Empirical Methods in Natural Language Processing. pp. 2383--2392. Association for Computational Linguistics, Austin, Texas (Nov 2016)

\bibitem{pwws}
Ren, S., Deng, Y., He, K., Che, W.: Generating natural language adversarial examples through probability weighted word saliency. In: Korhonen, A., Traum, D., M{\`a}rquez, L. (eds.) Proceedings of the 57th Annual Meeting of the Association for Computational Linguistics. pp. 1085--1097. Association for Computational Linguistics, Florence, Italy (Jul 2019)

\bibitem{multilingualbert}
Rosenthal, S., Bornea, M., Sil, A.: Are multilingual bert models robust? a case study on adversarial attacks for multilingual question answering. arXiv preprint arXiv:2104.07646  (2021)

\bibitem{DistilBERT}
Sanh, V., Debut, L., Chaumond, J., Wolf, T.: {DistilBERT}, a distilled version of {BERT}: smaller, faster, cheaper and lighter. In: Proceedings of the 5th Workshop on Energy Efficient Machine Learning and Cognitive Computing. NeurIPS, Vancouver, Canada (2019)

\bibitem{entroph-max}
Shinoda, K., Sugawara, S., Aizawa, A.: Penalizing confident predictions on largely perturbed inputs does not improve out-of-distribution generalization in question answering. In: Proceedings of the Workshop on Knowledge Augmented Methods for NLP (KnowledgeNLP) at AAAI 2023 (2023)

\bibitem{qa_review}
Soares, M.A.C., Parreiras, F.S.: A literature review on question answering techniques, paradigms and systems. Journal of King Saud University-Computer and Information Sciences  \textbf{32}(6),  635--646 (2020)

\bibitem{chatgpt}
Stiennon, N., Ouyang, L., Wu, J., Ziegler, D., Lowe, R., Voss, C., Radford, A., Amodei, D., Christiano, P.F.: Learning to summarize with human feedback. Advances in Neural Information Processing Systems  \textbf{33},  3008--3021 (2020)

\bibitem{sun2021adversarial}
Sun, H., Zhu, T., Zhang, Z., Jin, D., Xiong, P., Zhou, W.: Adversarial attacks against deep generative models on data: a survey. IEEE Transactions on Knowledge and Data Engineering  \textbf{35}(4),  3367--3388 (2021)

\bibitem{MultiQA}
Talmor, A., Berant, J.: {M}ulti{QA}: An empirical investigation of generalization and transfer in reading comprehension. In: Korhonen, A., Traum, D., M{\`a}rquez, L. (eds.) Proceedings of the 57th Annual Meeting of the Association for Computational Linguistics. pp. 4911--4921. Association for Computational Linguistics, Florence, Italy (Jul 2019)

\bibitem{tagger}
Toutanova, K., Klein, D., Manning, C.D., Singer, Y.: Feature-rich part-of-speech tagging with a cyclic dependency network. In: Proceedings of the 2003 Conference of the North American Chapter of the Association for Computational Linguistics on Human Language Technology - Volume 1. p. 173–180. NAACL '03, Association for Computational Linguistics, USA (2003)

\bibitem{NewsQA}
Trischler, A., Wang, T., Yuan, X., Harris, J., Sordoni, A., Bachman, P., Suleman, K.: {N}ews{QA}: A machine comprehension dataset. In: Blunsom, P., Bordes, A., Cho, K., Cohen, S., Dyer, C., Grefenstette, E., Hermann, K.M., Rimell, L., Weston, J., Yih, S. (eds.) Proceedings of the 2nd Workshop on Representation Learning for {NLP}. pp. 191--200. Association for Computational Linguistics, Vancouver, Canada (Aug 2017)

\bibitem{GBWI}
Wallace, E., Feng, S., Kandpal, N., Gardner, M., Singh, S.: Universal adversarial triggers for attacking and analyzing {NLP}. In: Proceedings of the 2019 Conference on Empirical Methods in Natural Language Processing and the 9th International Joint Conference on Natural Language Processing (EMNLP-IJCNLP). pp. 2153--2162. Association for Computational Linguistics, Hong Kong, China (November 2019)

\bibitem{wallace2019trick}
Wallace, E., Rodriguez, P., Feng, S., Yamada, I., Boyd-Graber, J.: Trick me if you can: Human-in-the-loop generation of adversarial examples for question answering. Transactions of the Association for Computational Linguistics  \textbf{7},  387--401 (2019)

\bibitem{t3}
Wang, B., Pei, H., Pan, B., Chen, Q., Wang, S., Li, B.: T3: Tree-autoencoder constrained adversarial text generation for targeted attack. In: Webber, B., Cohn, T., He, Y., Liu, Y. (eds.) Proceedings of the 2020 Conference on Empirical Methods in Natural Language Processing (EMNLP). pp. 6134--6150. Association for Computational Linguistics, Online (Nov 2020)

\bibitem{modernqa_survey}
Wang, Z.: Modern question answering datasets and benchmarks: A survey. arXiv preprint arXiv:2206.15030  (2022)

\bibitem{transformers}
Xiao, T., Zhu, J.: Introduction to transformers: an nlp perspective (2023)

\bibitem{show_att_tell}
Xu, K., Ba, J., Kiros, R., Cho, K., Courville, A., Salakhudinov, R., Zemel, R., Bengio, Y.: Show, attend and tell: Neural image caption generation with visual attention. In: International conference on machine learning. pp. 2048--2057. PMLR (2015)

\bibitem{att_road_sign}
Yang, X., Liu, W., Zhang, S., Liu, W., Tao, D.: Targeted attention attack on deep learning models in road sign recognition. IEEE Internet of Things Journal  \textbf{8}(6),  4980--4990 (2020)

\bibitem{robustmultilingual}
Yasunaga, M., Kasai, J., Radev, D.: Robust multilingual part-of-speech tagging via adversarial training. In: Walker, M., Ji, H., Stent, A. (eds.) Proceedings of the 2018 Conference of the North {A}merican Chapter of the Association for Computational Linguistics: Human Language Technologies, Volume 1 (Long Papers). pp. 976--986. Association for Computational Linguistics, New Orleans, Louisiana (Jun 2018)

\bibitem{qa_survey}
Yigit, G., Amasyali, M.F.: From text to multimodal: A comprehensive survey of adversarial example generation in question answering systems. Knowledge and Information Systems  \textbf{66},  7165--7204 (2024)

\bibitem{multihop}
Yu, J., Liu, W., Qiu, S., Su, Q., Wang, K., Quan, X., Yin, J.: Low-resource generation of multi-hop reasoning questions. In: Jurafsky, D., Chai, J., Schluter, N., Tetreault, J. (eds.) Proceedings of the 58th Annual Meeting of the Association for Computational Linguistics. pp. 6729--6739. Association for Computational Linguistics, Online (Jul 2020)

\bibitem{pso}
Zang, Y., Qi, F., Yang, C., Liu, Z., Zhang, M., Liu, Q., Sun, M.: Word-level textual adversarial attacking as combinatorial optimization. In: Jurafsky, D., Chai, J., Schluter, N., Tetreault, J. (eds.) Proceedings of the 58th Annual Meeting of the Association for Computational Linguistics. pp. 6066--6080. Association for Computational Linguistics, Online (Jul 2020)

\bibitem{RanMASK}
Zeng, J., Zheng, X., Xu, J., Li, L., Yuan, L., Huang, X.: Certified robustness to text adversarial attacks by randomized [mask]. Computational Linguistics  \textbf{49}(2),  395--427 (jun 2023)

\bibitem{pegasus}
Zhang, J., Zhao, Y., Saleh, M., Liu, P.: Pegasus: Pre-training with extracted gap-sentences for abstractive summarization. In: International conference on machine learning. pp. 11328--11339. PMLR (2020)

\bibitem{search_engine}
Zhu, F., Lei, W., Wang, C., Zheng, J., Poria, S., Chua, T.S.: Retrieving and reading: A comprehensive survey on open-domain question answering. arXiv preprint arXiv:2101.00774  (2021)

\bibitem{RoBERTa}
Zhuang, L., Wayne, L., Ya, S., Jun, Z.: A robustly optimized {BERT} pre-training approach with post-training. In: Li, S., Sun, M., Liu, Y., Wu, H., Liu, K., Che, W., He, S., Rao, G. (eds.) Proceedings of the 20th Chinese National Conference on Computational Linguistics. pp. 1218--1227. Chinese Information Processing Society of China, Huhhot, China (Aug 2021)

\end{thebibliography}
\end{document}